\documentclass[letterpaper, 10 pt, journal, twoside]{IEEEtran}

\IEEEoverridecommandlockouts                              % This command is only needed if 
\usepackage{amsmath} % assumes amsmath package installed
\usepackage{amssymb}  % assumes amsmath package installed

\usepackage[dvipdfmx]{graphicx}
\usepackage{url}
\usepackage{multirow}
\usepackage{siunitx}

\newcommand{\figref}[1]{{Fig.~\ref{#1}}}

\newcommand{\bm}[1]{\mbox{\boldmath{$#1$}}}

\title{Centroidal Trajectory Generation and Stabilization based on Preview Control for Humanoid Multi-contact Motion}

\author{Masaki Murooka, Mitsuharu Morisawa and Fumio Kanehiro% <-this % stops a space
  \thanks{Manuscript received: February, 16, 2022; Revised May, 9, 2022; Accepted June, 23, 2022.}%Use only for final RAL version
  \thanks{This paper was recommended for publication by Editor L. Pallottino upon evaluation of the Associate Editor and Reviewers' comments.
  This work was supported by KAKENHI (21H01300).} %Use only for final RAL version
  \thanks{The authors are with
    CNRS-AIST JRL (Joint Robotics Laboratory), IRL and
    National Institute of Advanced Industrial Science and Technology (AIST),
    1-1-1 Umezono, Tsukuba, Ibaraki 305-8560, Japan.
    {\tt\small \{m-murooka, m.morisawa, f-kanehiro\}@aist.go.jp}}%
  \thanks{Digital Object Identifier (DOI): see top of this page.}
}

\begin{document}

\maketitle

\markboth{IEEE Robotics and Automation Letters. Preprint Version. Accepted June, 2022}
{Murooka \MakeLowercase{\textit{et al.}}: Centroidal Trajectory Generation and Stabilization based on Preview Control}

%% floatsep is 12.0pt plus 2.39996pt minus 2.39996pt (= 12.0pt)
%% textfloatsep is 20.39996pt plus 2.39996pt minus 4.79993pt (= 18.0pt)
%% abovecaptionskip is 6.0pt
%% abovedisplayskip is 6.74997pt plus 4.0pt minus 2.0pt (= 8.74997pt)
%% belowdisplayskip is 6.74997pt plus 4.0pt minus 2.0pt (= 8.74997pt)
\setlength{\floatsep}{10pt}
\setlength{\textfloatsep}{16pt}
\setlength{\abovecaptionskip}{4pt}
\setlength{\abovedisplayskip}{5pt}
\setlength{\belowdisplayskip}{5pt}

%%%%%%%%%%%%%%%%%%%%%%%%%%%%%%%%%%%%%%%%%%%%%%%%%%%%%%%%%%%%%%%%%%%%%%%%%%%%%%%%
\begin{abstract}
  Multi-contact motion is important for humanoid robots to work in various environments.
  We propose a centroidal online trajectory generation and stabilization control for humanoid dynamic multi-contact motion.
  The proposed method features the drastic reduction of the computational cost by using preview control instead of the conventional model predictive control that considers the constraints of all sample times.
  By combining preview control with centroidal state feedback for robustness to disturbances and wrench distribution for satisfying contact constraints, we show that the robot can stably perform a variety of multi-contact motions through simulation experiments.
\end{abstract}


%%%%%%%%%%%%%%%%%%%%%%%%%%%%%%%%%%%%%%%%%%%%%%%%%%%%%%%%%%%%%%%%%%%%%%%%%%%%%%%%
\section{Introduction}

\IEEEPARstart{H}{umanoid} robots can perform locomotion and manipulation that maximize the potential of the high-degree-of-freedom body through multi-contact motion using not only the feet but also the hands.
The planning and control of such multi-contact motion have been studied extensively in the last decade.
In particular, many studies have focused on the robot centroidal dynamics~\cite{CentroidalDynamics:Orin:AuRo2013} instead of the full dynamics, thereby reducing the dimensionality and computational cost~\cite{WholebodyPlanning:Dai:Humanoids2014,MultiContact:Carpentier:TRO2018,CentroidalDynamicsBCD:Shah:IROS2021}.
However, even focusing on the centroidal dynamics, it is still non-trivial to realize receding horizon control that can be executed on the order of submillisecond due to its nonlinearity and contact constraints.

In this paper, we propose the centroidal online trajectory generation method based on preview control and the stabilization control method to compensate for the error of the centroidal state.
The proposed method takes a new approach that is entirely different from model predictive control (MPC), which explicitly considers time series constraints~\cite{MultiContact:Audren:IROS2014,HumanoidMPC:Henze:IROS2014,MultiContactMPC:Caron:Humanoids2016,CROC:Fernbach:IROS2018,ContactAdjustmentNMPC:Romualdi:ICRA2022}, and has the advantage of significantly lowering computational costs.
In the simulation experiments of humanoid multi-contact motion, our method can sequentially generate the centroidal trajectory with stabilization control within $1$~ms, considering the reference input of the horizon of 2~s (400 sample points); to the best of our knowledge, this is one of the fastest centroidal trajectory generation and control methods that can handle general multi-contact motion without relying on the biped-specific dynamics.
The effectiveness of the proposed method is demonstrated through simulation experiments in which a humanoid robot performs various challenging dynamic multi-contact motions.

%% \begin{figure}[tpb]
%%   \begin{center}
%%     \includegraphics[width=0.5\columnwidth]{figs/nowprinting.eps}
%%     \caption{TODO.
%%     }
%%     \label{fig:intro}
%%   \end{center}
%% \end{figure}

%%%%%%%%
\subsection{Related Works}

\subsubsection{Centroidal Online Trajectory Generation}

By generating a centroidal trajectory with the receding horizon control scheme, the robot can robustly respond to environmental changes and tracking errors by flexibly modifying the position and timing of contact transitions~\cite{SequentialMultiContact:Morisawa:IJHR2020}.
MPC, which is formulated as a quadratic programming (QP) problem, has been widely used for the receding horizon control of legged robots~\cite{MultiContact:Audren:IROS2014,HumanoidMPC:Henze:IROS2014,MultiContactMPC:Caron:Humanoids2016,CROC:Fernbach:IROS2018,ContactAdjustmentNMPC:Romualdi:ICRA2022}.
In MPC, equality and inequality constraints including contact force constraints are imposed on all input / state variables of sampling times in the horizon.
Therefore, the dimensionality of the constraints tends to be large, and it is not easy to solve it on the order of submillisecond, even with state-of-the-art QP solvers.

In this paper, we propose an application of preview control~\cite{PreviewControl:Katayama:IJC1985}, which does not explicitly consider these constraints.
Since the preview control requires only a single matrix multiplication at runtime and does not require an iterative computation, it is considerably faster than MPC, which requires solving a QP.
We show that even though the constraints are omitted in the preview control, a realistic reference trajectory (but a fairly rough one that is discontinuous and non-differentiable) and post-processing wrench projection can generate feasible robot motions.
We note that differential dynamic programming, which has recently been used for robot motion generation~\cite{Crocoddyl:Mastalli:ICRA2020}, can also be applied to fast unconstrained receding horizon control.

\subsubsection{Balance Stabilization Control}

Our proposed centroidal stabilization control is similar to the balance control~\cite{HumanoidBalance:Ott:Humanoids2011} since it provides feedback on the centroidal position and orientation.
However, our method differs in that the resultant wrench obtained from online trajectory generation is used for feedforward.
While some stabilization control methods for multi-contact motion target specific tasks (e.g., ladder climbing~\cite{RobustLadder:Kanazawa:IROS2015} or whole-body pushing~\cite{MultiContactPush:Polverini:RAL2020}), our method can handle various multi-contact motions including locomotion on flat ground, handrail stairs, and vertical ladders.

For bipedal walking, stabilization control based on the divergent component of motion (DCM) has been widely proposed~\cite{DcmWalk:Englsberger:TRO2015,StairClimb:Caron:ICRA2019}.
Although this is a powerful method, it relies on the linear inverted pendulum mode (LIPM)~\cite{LIPM:Kajita:IROS2001}, which assumes bipedal dynamics, and is difficult to extend to general multi-contact motion.
In this paper, we show that the proposed centroidal control for multi-contact motion includes the proportional control of DCM for bipedal walking as a particular case, and derive the relationship between the gain parameters of the two controls.

\begin{figure*}[tpb]
  \begin{center}
    \includegraphics[width=2.0\columnwidth]{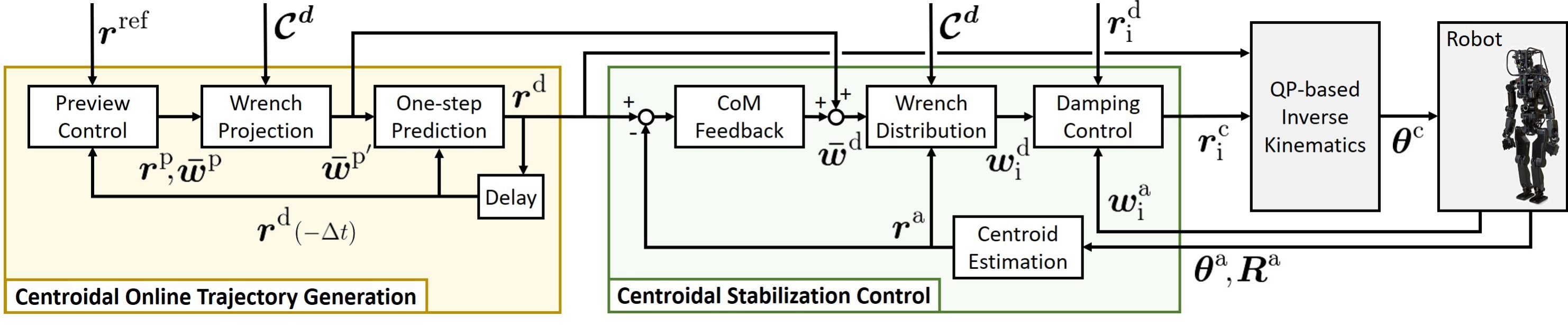}
    \caption{Overall components of the control system for humanoid multi-contact motion.
    }
    \label{fig:system}
  \end{center}
\end{figure*}

%%%%%%%%
\subsection{Contributions of this Paper}

The contributions of our work are threefold:
(i) online generation method of robot centroidal trajectory with very low computational cost based on preview control,
(ii) a control system that can handle humanoid bipedal walking and multi-contact motion including hand contacts without changing the parameters,
(iii) showing that a humanoid robot can perform various challenging multi-contact motions such as handrail stairs and vertical ladders in simulation.

%%%%%%%%%%%%%%%%%%%%%%%%%%%%%%%%%%%%%%%%%%%%%%%%%%%%%%%%%%%%%%%%%%%%%%%%%%%%%%%%
\section{Method Overview} \label{sec:system}

\figref{fig:system} shows the proposed control system for humanoid multi-contact motion.
The control system consists mainly of centroidal online trajectory generation and stabilization control.
These two modules are introduced in Sections~\ref{sec:plan} and \ref{sec:control}, respectively, followed by simulation experiments in Section~\ref{sec:exp}.
%% We present these two modules in Sections~\ref{sec:plan} and \ref{sec:control}, respectively.
%% Then, simulation experiments are shown in Section~\ref{sec:exp}, and the proposed method is supplemented in Section ??, including its relation to bipedal walking control.

We assume that the robot is joint position controlled.
The sequence of contact positions $\bm{\mathcal{C}}^{\mathrm{d}}$ of the hands and feet is assumed to be determined by a global planner~\cite{MultiContactSurvey:Bouyarmane:Reference2018}, or manually.

%%%%%%%%%%%%%%%%%%%%%%%%%%%%%%%%%%%%%%%%%%%%%%%%%%%%%%%%%%%%%%%%%%%%%%%%%%%%%%%%
\section{Centroidal Online Trajectory Generation} \label{sec:plan}

%%%%%%%%
\subsection{6-DoF Centroidal Preview Control}

\subsubsection{Approximated Centroidal Dynamics}

The centroidal dynamics of a humanoid robot are governed by the Newton-Euler equation~\cite{ComEstimate:Carpentier:TRO2016}:
\begin{subequations} \label{eq:centroidal-dynamics}
\begin{align}
  m \bm{\ddot{c}} &= \bm{f} - m \bm{g} \label{eq:centroidal-dynamics-linear} \\
  \bm{\dot{L}} &= \bm{n} - \bm{c} \times \bm{f} \label{eq:centroidal-dynamics-angular}
\end{align}
\end{subequations}
$m \in \mathbb{R}$ is the robot mass.
$\bm{c} \in \mathbb{R}^3$ is the robot CoM.
$\bm{L} \in \mathbb{R}^3$ is the centroidal angular momentum around the CoM.
$\bm{g} = [0 \ 0 \ g]^{\mathsf{T}}$ is the (vertical upward) gravitational acceleration vector.
$\bm{f}, \bm{n} \in \mathbb{R}^3$ are the force and moment from the contacts between the robot and the environment, represented in world coordinates.
Especially, $\bm{n}$ is the moment around the world origin.

We introduce the resultant force and moment, which combines the effects of contacts and gravity, as follows:
\begin{subequations} \label{eq:wrench-without-gravity}
\begin{align}
  \bm{\bar{f}} &= \bm{f} - m \bm{g} \label{eq:wrench-without-gravity-linear} \\
  \bm{\bar{n}} &= \bm{n} - \bm{c} \times \bm{f} \label{eq:wrench-without-gravity-angular}
\end{align}
\end{subequations}
In the following, the symbols with bar denote the resultant force and moment combined with gravity.
%% Combining \eqref{eq:centroidal-dynamics} and \eqref{eq:wrench-without-gravity}, we obtain the following:
%% %% Assuming that the second term on the left-hand side of \eqref{eq:centroidal-dynamics-angular} can be ignored~\cite{HumanoidMPC:Henze:IROS2014,LeggedControl:Sombolestan:IROS2021}, \eqref{eq:centroidal-dynamics} becomes as follows:
%% \begin{subequations} \label{eq:centroidal-dynamics-approx}
%% \begin{align}
%%   m \bm{\ddot{c}} &= \bm{\bar{f}} \label{eq:centroidal-dynamics-approx-linear} \\
%%   \bm{\dot{L}} &= \bm{\bar{n}} \label{eq:centroidal-dynamics-approx-angular}
%% \end{align}
%% \end{subequations}

%%
\subsubsection{State Equation for Preview Control}

The linear components of centroidal dynamics \eqref{eq:centroidal-dynamics-linear} are represented by the following state equations:
\begin{subequations} \label{eq:pc-state-eq-linear}
\begin{align}
  \bm{\dot{x}}_{\mathrm{L*}} &=
  \begin{bmatrix} 0 & 1 & 0 \\ 0 & 0 & 1 \\ 0 & 0 & 0 \end{bmatrix}
  \bm{x}_{\mathrm{L*}}
  +
  \begin{bmatrix} 0 \\ 0 \\ 1 \end{bmatrix}
  u_{\mathrm{L*}}
  \\
  \bm{y}_{\mathrm{L*}} &=
  \begin{bmatrix} 1 & 0 & 0 \\ 0 & 0 & m \end{bmatrix}
  \bm{x}_{\mathrm{L*}} \\
  & {\rm where} \ \ \ \ \bm{x}_{\mathrm{L*}} = \begin{bmatrix} c_* & \dot{c}_* & \ddot{c}_* \end{bmatrix}^{\mathsf{T}} \nonumber \\
  & \phantom{\rm where} \ \ \ \ u_{\mathrm{L*}} = \dddot{c}_* \nonumber \\
  & \phantom{\rm where} \ \ \ \ \bm{y}_{\mathrm{L*}} = \begin{bmatrix} c_* & \bar{f}_* \end{bmatrix}^{\mathsf{T}} \nonumber \\
  & \phantom{\rm where} \ \ \ \ * \in \{x, y, z\} \nonumber
\end{align}
\end{subequations}
$c_*$ and $\bar{f}_* \ (* \in \{x, y, z\})$ represent the axis components of $\bm{c}$ and $\bm{\bar{f}}$, respectively.
The states consist of the CoM position, velocity, and acceleration.
The input is the CoM jerk.
The outputs consist of the CoM and the resultant force.

Compared to the preview control for bipedal walking by Kajita et al.~\cite{PreviewControl:Kajita:ICRA2003}, the definitions of states and input are the same, but the definition of output is different.
In bipedal walking~\cite{PreviewControl:Kajita:ICRA2003}, the ZMP is regarded as an output to follow the reference ZMP trajectory, whereas in our method, the CoM and the force are regarded as outputs to support general multi-contact motion.
We chose this output definition from the consideration that the ZMP in bipedal walking has force information as the center of pressure and also has CoM information based on the CoM-ZMP relationship.
One of the main contributions of this paper is that preview control with this output definition works effectively to generate general multi-contact motions.

\subsubsection{Optimal Input Calculation}

Given reference trajectories for the robot CoM and resultant force, the following objective function is minimized:
\begin{subequations} \label{eq:pc-objective-linear}
\begin{align}
  J_{\mathrm{L*}} &= \sum_{i = k}^{\infty} \left(
  \left\| \bm{y}_{\mathrm{L*}}[i] - \bm{y}_{\mathrm{L*}}^{\mathrm{ref}}[i] \right\|^2_{\!_{\bm{Q}_{\mathrm{L}}}} + \left\| u_{\mathrm{L*}}[i] \right\|^2_{\!_{\bm{R}_{\mathrm{L}}}}
  \right) \\
  &= \sum_{i = k}^{\infty} \left( \left\| \begin{bmatrix} c_*[i] - c_*^{\mathrm{ref}}[i] \\ \bar{f}_*[i] - \bar{f}_*^{\mathrm{ref}}[i] \end{bmatrix} \right\|^2_{\!_{\bm{Q}_{\mathrm{L}}}} + \left\| \dddot{c}_*[i] \right\|^2_{\!_{\bm{R}_{\mathrm{L}}}} \right)
\end{align}
\end{subequations}
$\bm{Q}_{\mathrm{L}}$ and $\bm{R}_{\mathrm{L}}$ are the objective weights of output and input, respectively.
For a general vector $\bm{x}$, $\left\|\bm{x}\right\|^2_{\scriptsize \bm{W}}$ represents $\bm{x}^{\mathsf{T}} \bm{W} \bm{x}$ where $\bm{W}$ is $\bm{Q}_{\mathrm{L}}$ or $\bm{R}_{\mathrm{L}}$.
$k$ is the control step index.
We always set the reference force to zero.
The reference CoM trajectory was determined by simple rules such as the center of the contact points.
See Section~\ref{sec:exp} for examples.

According to the preview control theory~\cite{PreviewControl:Katayama:IJC1985}, the optimal input can be obtained as follows:
\begin{align}
  u^{\mathrm{opt}}_{\mathrm{L*}}[k] = - \bm{K}_{\mathrm{fb}} \bm{x}_{\mathrm{L*}}[k] + \sum_{i = 1}^{N_h} \bm{K}_{\mathrm{ff}}[i] \; \bm{y}_{\mathrm{L*}}^{\mathrm{ref}} [k+i] \label{eq:pc-calc-input-linear}
\end{align}
$N_h$ is the number of time-steps of the preview window.
Since the state equation \eqref{eq:pc-state-eq-linear} depends only on the robot mass, the gains $\bm{K}_{\mathrm{fb}}$ and $\bm{K}_{\mathrm{ff}}$ are fixed as long as the robot mass is constant.
See \cite{PreviewControl:Katayama:IJC1985} and \cite{PreviewControl:Kajita:ICRA2003} for the derivation of these gain matrices.
The equation \eqref{eq:pc-calc-input-linear} requires only matrix multiplication, which is considerably less computationally expensive than conventional multi-contact motion generation methods~\cite{MultiContactSurvey:Bouyarmane:Reference2018} such as MPC with constraints~\cite{MultiContact:Audren:IROS2014} or trajectory optimization~\cite{WholebodyPlanning:Dai:Humanoids2014}.
Although our preview control does not consider constraints on contact forces, we show in Section~\ref{sec:exp} that a realistic reference trajectory (but a fairly rough one) and post-processing can produce feasible robot motions.

\subsubsection{6-DoF Trajectory Generation}

We approximate the centroidal angular momentum in \eqref{eq:centroidal-dynamics-angular} as follows:
\begin{align}
  \bm{L} = \bm{I}_{\mathrm{all}} \, \bm{\dot{q}} \approx \bm{I}_{\mathrm{base}} \, \bm{\omega} \approx \bm{I} \bm{\omega} \label{eq:angular-momentum-approx}
\end{align}
$\bm{\dot{q}} \in \mathbb{R}^{N_{\dot{q}}}$ is the joint velocity vector including the 6-DoF virtual joint velocity for the robot base link.
$\bm{I}_{\mathrm{all}} \in \mathbb{R}^{3 \times N_{\dot{q}}}$ is the angular part of the centroidal momentum matrix~\cite{CentroidalDynamics:Orin:AuRo2013}.
$\bm{\omega} \in \mathbb{R}^3$ is the angular velocity of the base link.
$\bm{I}_{\mathrm{base}} \in \mathbb{R}^{3 \times 3}$ is the block matrix of $\bm{I}_{\mathrm{all}}$ corresponding to the base link angular velocity.
$\bm{I} \in \mathbb{R}^{3 \times 3}$ is a constant and diagonal approximation of $\bm{I}_{\mathrm{base}}$.
The approximation in equation \eqref{eq:angular-momentum-approx} assumes that
(i) the angular momentum due to joint velocity is negligible, (ii) $\bm{I}_{\mathrm{base}}$ is independent of joint position and base link orientation, and (iii) off-diagonal elements of $\bm{I}_{\mathrm{base}}$ are negligible.
The validity of these strong assumptions in humanoid multi-contact motion is evaluated in Section~\ref{sec:val-rot-motion}.
%% We further assume that the inertial tensor is independent for each component; that is, $\bm{I}$ is a diagonal matrix.
%% Even under these strong assumptions, we show that it works effectively in the dynamic multi-contact motion of a humanoid robot in simulation by integrating with stabilization control.
%% We also discuss in Section ?? how to relax the limitations of these assumptions.

Similar to the linear components, the angular components of centroidal dynamics \eqref{eq:centroidal-dynamics-angular}, under the approximation \eqref{eq:angular-momentum-approx}, are represented by the following state equations:
\begin{subequations} \label{eq:pc-state-eq-angular}
\begin{align}
  \bm{\dot{x}}_{\mathrm{A*}} &=
  \begin{bmatrix} 0 & 1 & 0 \\ 0 & 0 & 1 \\ 0 & 0 & 0 \end{bmatrix}
  \bm{x}_{\mathrm{A*}}
  +
  \begin{bmatrix} 0 \\ 0 \\ 1 \end{bmatrix}
  u_{\mathrm{A*}}
  \label{eq:pc-state-eq-angular-1} \\
  \bm{y}_{\mathrm{A*}} &=
  \begin{bmatrix} 1 & 0 & 0 \\ 0 & 0 & I_* \end{bmatrix}
  \bm{x}_{\mathrm{A*}} \\
  & {\rm where} \ \ \ \ \bm{x}_{\mathrm{A*}} = \begin{bmatrix} \alpha_* & \dot{\alpha}_* & \ddot{\alpha}_* \end{bmatrix}^{\mathsf{T}} \nonumber \\
  & \phantom{\rm where} \ \ \ \ u_{\mathrm{A*}} = \dddot{\alpha}_* \nonumber \\
  & \phantom{\rm where} \ \ \ \ \bm{y}_{\mathrm{A*}} = \begin{bmatrix} \alpha_* & \bar{n}_* \end{bmatrix}^{\mathsf{T}} \nonumber \\
  & \phantom{\rm where} \ \ \ \ * \in \{x, y, z\} \nonumber
\end{align}
\end{subequations}
$\bm{\alpha} \in \mathbb{R}^3$ is the Euler angle representing the base link orientation.
$\alpha_*$ and $\bar{n}_* \ (* \in \{x, y, z\})$ represent the axis components of $\bm{\alpha}$ and $\bm{\bar{n}}$, respectively.
$I_*$ is the diagonal element of $\bm{I}$.
Although the relation $\bm{\omega} = \bm{K}_{\mathrm{Euler}}(\bm{\alpha}) \, \bm{\dot{\alpha}}$ holds, $\bm{K}_{\mathrm{Euler}}$ is ignored in \eqref{eq:pc-state-eq-angular}.
Its validity is evaluated in Section~\ref{sec:val-rot-motion}.
%% In \eqref{eq:pc-state-eq-angular-1}, each component of the Euler angle is differentiated independently as an approximation.
%% We discuss the relaxation of this approximation in Section ??.
Compared to the preview control for bipedal walking~\cite{PreviewControl:Kajita:ICRA2003}, the proposed method's prominent feature is generating a 6-DoF centroidal trajectory, including position and orientation.

The optimal inputs $u^{\mathrm{opt}}_{\mathrm{A*}}$ for the same objective function as \eqref{eq:pc-objective-linear} can be obtained by preview control in the same way as \eqref{eq:pc-calc-input-linear}.

From the optimal inputs $\bm{\dddot{c}}^{\mathrm{opt}}, \bm{\dddot{\alpha}}^{\mathrm{opt}}$,
we obtain the planned centroidal state $\bm{r}^{\mathrm{p}} = [{\bm{c}^{\mathrm{p}}}^{\mathsf{T}} \ {\bm{\alpha}^{\mathrm{p}}}^{\mathsf{T}}]^{\mathsf{T}}$ and the planned resultant wrench $\bm{\bar{w}}^{\mathrm{p}} = [{\bm{\bar{f}}^{\mathrm{p}}}^{\mathsf{T}}, \ {\bm{\bar{n}}^{\mathrm{p}}}^{\mathsf{T}}]^{\mathsf{T}}$
by state equations \eqref{eq:pc-state-eq-linear} and \eqref{eq:pc-state-eq-angular}, respectively.

%%%%%%%%
\subsection{Resultant Wrench Projection}

\subsubsection{Wrench Distribution Formulation}

\begin{figure}[tpb]
  \begin{center}
    \includegraphics[width=0.9\columnwidth]{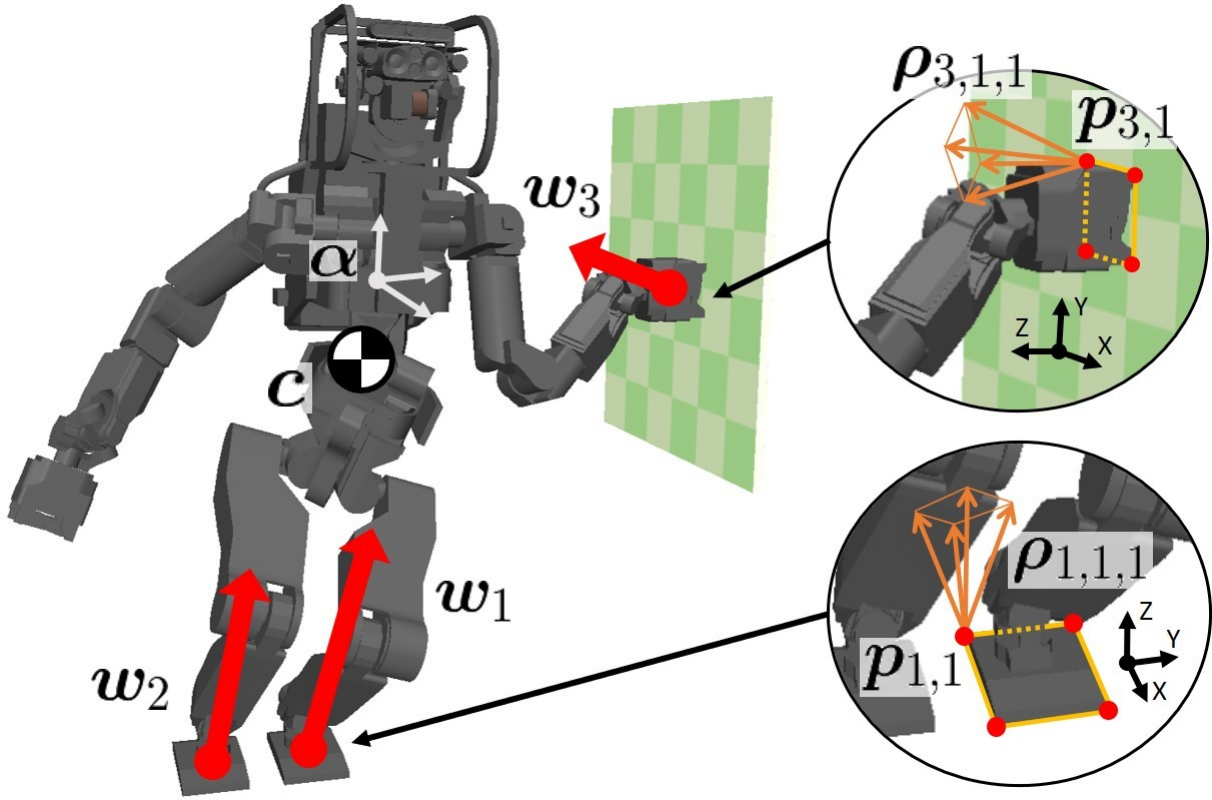}
    \caption{Contact constraints in wrench distribution.
    }
    \label{fig:wrench-distrib}
  \end{center}
\end{figure}

The resultant wrench $\bm{\bar{w}}^{\mathrm{p}}$ planned by preview control is not always feasible because the forces the robot receives from the environment are subject to contact constraints.
We modify the resultant wrench to be feasible by projecting it onto the constraint manifold defined by the contacts.

The contact wrench at the i-th limb end $\bm{w}_{\mathrm{i}} = [\bm{f}_{\mathrm{i}}^{\mathsf{T}} \ \bm{n}_{\mathrm{i}}^{\mathsf{T}}]^{\mathsf{T}}$ is represented as follows~\cite{MultiContactSurvey:Bouyarmane:Reference2018}:
%% \begin{subequations}
\begin{align}
  \bm{w}_{\mathrm{i}}
  &=
  \bm{G}_{\mathrm{i}} \bm{\lambda}_{\mathrm{i}} \\
  & {\rm where} \ \ \ \
  \bm{G}_{\mathrm{i}} =
  \begin{bmatrix}
    \cdots & \bm{\rho}_{\mathrm{i,j,k}} & \cdots \\
    \cdots & \bm{p}_{\mathrm{i,j}} \! \times \! \bm{\rho}_{\mathrm{i,j,k}} & \cdots
  \end{bmatrix} \nonumber \\
  & \phantom{\rm where} \ \ \ \
  \bm{\lambda}_{\mathrm{i}} =
  \begin{bmatrix}
    \vdots \\ \lambda_{\mathrm{i,j,k}} \\ \vdots
  \end{bmatrix}, \ \ \bm{\lambda}_{\mathrm{i}} \geq \bm{0} \nonumber
\end{align}
%% \end{subequations}
$\bm{p}_{\mathrm{i,j}} \in \mathbb{R}^3 (j = 1,\cdots,J(i))$ is the j-th vertex of the contact polygon.
$\bm{\rho}_{\mathrm{i,j,k}} \in \mathbb{R}^3 (k = 1,\cdots,K(i,j))$ is the k-th ridge vector of the friction pyramid at the j-th vertex.
\figref{fig:wrench-distrib} illustrates an example of these variables.
All variables are represented in world coordinates.
The grasping contacts (e.g., the hand contacts in \figref{fig:exp-multi-contact}~(C)) are represented by placing the friction pyramids facing each other.

\subsubsection{Wrench Projection}

The contact wrench required to achieve the resultant wrench closest to the planned resultant wrench can be obtained by solving the following QP problem\footnote{To direct the contact force away from the friction cone boundary, a penalty term weighted in the local contact coordinates can be added to the objective function~\cite{QuadrupedSlope:Focchi:AuRo2017}.}:
%% \begin{subequations}
\begin{align} \label{eq:wrench-distrib-qp-plan}
  & \min_{\scriptsize \bm{\lambda}} \ \ \| \bm{G} \bm{\lambda} - \bm{w}^{\mathrm{p}} \|^2 \ \ \ \ {\rm s.t.} \ \ \ \ \bm{\lambda} \geq \bm{0} \\
  & {\rm where} \ \ \ \ \bm{G} = \begin{bmatrix} \cdots & \bm{G}_{\mathrm{i}} & \cdots \end{bmatrix} \nonumber \\
  & \phantom{\rm where} \ \ \ \ \bm{\lambda} = \begin{bmatrix} \cdots & \bm{\lambda}_{\mathrm{i}}^{\mathsf{T}} & \cdots \end{bmatrix}^{\mathsf{T}} \nonumber \\
  & \phantom{\rm where} \ \ \ \ \bm{w}^{\mathrm{p}} = \bm{\bar{w}}^{\mathrm{p}} + \begin{bmatrix} m \bm{g} \\ \bm{c}^{\mathrm{p}} \times \bm{f}^{\mathrm{p}} \end{bmatrix} \nonumber
\end{align}
%% \end{subequations}
$\bm{G}$ is calculated from the target contact state $\bm{\mathcal{C}}^{\mathrm{d}}$, which is treated as given, including the contact polygons and friction coefficients.

With $\bm{\lambda}^{\mathrm{opt}}$ as the optimal variable in \eqref{eq:wrench-distrib-qp-plan}, the planned resultant wrench projected onto the contact constraint manifold is calculated as follows:
\begin{align}
  \bm{\bar{w}}^{\mathrm{p}^\prime} = \bm{G} \bm{\lambda}^{\mathrm{opt}} - \begin{bmatrix} m \bm{g} \\ \bm{c}^{\mathrm{p}} \times \bm{f}^{\mathrm{p}} \end{bmatrix}
\end{align}

By inputting $\bm{\bar{w}}^{\mathrm{p}^\prime}$ and integrating \eqref{eq:centroidal-dynamics} for one control period, the desired centroidal state of the next control loop $\bm{r}^{\mathrm{d}} = [{\bm{c}^{\mathrm{d}}}^{\mathsf{T}} \ {\bm{\alpha}^{\mathrm{d}}}^{\mathsf{T}}]^{\mathsf{T}}$ is obtained.

%%%%%%%%%%%%%%%%%%%%%%%%%%%%%%%%%%%%%%%%%%%%%%%%%%%%%%%%%%%%%%%%%%%%%%%%%%%%%%%%
\section{Centroidal Stabilization Control} \label{sec:control}

%%%%%%%%
\subsection{Centroidal Feedback Control}

The stabilizer is responsible for reducing the error between the desired centroidal state $\bm{r}^{\mathrm{d}}$ and the actual centroidal state $\bm{r}^{\mathrm{a}}$ based on sensor measurements.
We determine the modification amount of the resultant wrench by PD control of the centroidal state as follows:
\begin{subequations} \label{eq:com-feedback}
\begin{align}
  \Delta \bm{\bar{w}}^{\mathrm{d}} &= \bm{K}_{\mathrm{P}} (\bm{r}^{\mathrm{d}} - \bm{r}^{\mathrm{a}}) + \bm{K}_{\mathrm{D}} (\bm{\dot{r}}^{\mathrm{d}} - \bm{\dot{r}}^{\mathrm{a}}) \\
  &= \begin{bmatrix}
    \bm{K}_{\mathrm{P_L}} (\bm{c}^{\mathrm{d}} - \bm{c}^{\mathrm{a}}) + \bm{K}_{\mathrm{D_L}} (\bm{\dot{c}}^{\mathrm{d}} - \bm{\dot{c}}^{\mathrm{a}}) \\
    \bm{K}_{\mathrm{P_A}} \mathrm{log} \left( \bm{R}^{\mathrm{d}} {\bm{R}^{\mathrm{a}}}^{\mathsf{T}} \right) + \bm{K}_{\mathrm{D_A}} (\bm{\dot{\alpha}}^{\mathrm{d}} - \bm{\dot{\alpha}}^{\mathrm{a}}) \label{eq:com-feedback-2}
  \end{bmatrix}
\end{align}
\end{subequations}
$\bm{R}^{\mathrm{d}}$ and $\bm{R}^{\mathrm{a}}$ are the rotation matrices corresponding to $\bm{\alpha}^{\mathrm{d}}$ and $\bm{\alpha}^{\mathrm{a}}$, respectively.
$\mathrm{log}(\bm{R}) \in \mathbb{R}^3$ is a function that converts the rotation matrix $\bm{R}$ to an equivalent axis-angle vector.
Then, the desired resultant wrench is calculated as follows:
\begin{align}
  \bm{\bar{w}}^{\mathrm{d}} = \bm{\bar{w}}^{\mathrm{p}^\prime} + \Delta \bm{\bar{w}}^{\mathrm{d}}
\end{align}
$\bm{K}_{\mathrm{P}}, \bm{K}_{\mathrm{D}} \in \mathbb{R}^{6 \times 6}$ are the diagonal matrices of the feedback gains.
We describe the similarity between this control and the bipedal walking control based on the DCM in Appendix-A.

%%%%%%%%
\subsection{Resultant Wrench Distribution}

The desired resultant wrench is distributed to the contact wrenches at the limb ends by solving the following QP problem:
%% \begin{subequations}
\begin{align} \label{eq:wrench-distrib-qp-control}
  & \min_{\scriptsize \bm{\lambda}} \ \ \| \bm{G} \bm{\lambda} - \bm{w}^{\mathrm{d}} \|^2 \ \ \ \ {\rm s.t.} \ \ \ \ \bm{\lambda} \geq \bm{0} \\
  & {\rm where} \ \ \ \ \bm{w}^{\mathrm{d}} = \bm{\bar{w}}^{\mathrm{d}} + \begin{bmatrix} m \bm{g} \\ \bm{c}^{\mathrm{a}} \times \bm{f}^{\mathrm{d}} \end{bmatrix} \nonumber
\end{align}
%% \end{subequations}
The definitions of $\bm{G}$ and $\bm{\lambda}$ are the same as in \eqref{eq:wrench-distrib-qp-plan}.
From the experimental results in the simulation, it was confirmed that the stability performance is improved by using $\bm{c}^{\mathrm{a}}$ instead of $\bm{c}^{\mathrm{d}}$.

With $\bm{\lambda}^{\mathrm{opt}}$ as the optimal variable in \eqref{eq:wrench-distrib-qp-control}, the desired contact wrench of the i-th limb end is calculated as follows:
\begin{align}
  \bm{w}^{\mathrm{d}}_{\mathrm{i}} = \bm{G}_{\mathrm{i}} \bm{\lambda}_{\mathrm{i}}^{\mathrm{opt}}
\end{align}

%%%%%%%%
\subsection{Damping Control for Limb Ends}

Damping control~\cite{Stabilizer:Kajita:IROS2010} is applied to achieve the desired contact wrench $\bm{w}^{\mathrm{d}}_{\mathrm{i}}$ of each limb end.

Let $\bm{p}_{\mathrm{i}}^{\mathrm{d}} \in \mathbb{R}^3$ and $\bm{R}_{\mathrm{i}}^{\mathrm{d}} \in \mathbb{R}^{3 \times 3}$ represent the desired pose of the limb end determined from the given target contact position, and $\bm{p}_{\mathrm{i}}^{\mathrm{c}}$ and $\bm{R}_{\mathrm{i}}^{\mathrm{c}}$ represent the compliance pose of the limb end.
In damping control, the compliance pose is updated to satisfy the following relationship:
\begin{align}
  & \bm{K}_{\mathrm{d}} \Delta \bm{\dot{r}}_{\mathrm{i}}^{\mathrm{c}} + \bm{K}_{\mathrm{s}} \Delta \bm{r}_{\mathrm{i}}^{\mathrm{c}} = \bm{K}_{\mathrm{f}} (\bm{w}_{\mathrm{i}}^{\mathrm{a}} - \bm{w}_{\mathrm{i}}^{\mathrm{d}}) \label{eq:damping-control} \\
  & {\rm where} \ \ \ \ \Delta \bm{r}_{\mathrm{i}}^{\mathrm{c}} =
  \begin{bmatrix} \Delta \bm{r}_{\mathrm{i,L}}^{\mathrm{c}} \\ \Delta \bm{r}_{\mathrm{i,A}}^{\mathrm{c}} \end{bmatrix} =
  \begin{bmatrix} \bm{p}_{\mathrm{i}}^{\mathrm{c}} - \bm{p}_{\mathrm{i}}^{\mathrm{d}} \\ \mathrm{log}\left(\bm{R}_{\mathrm{i}}^{\mathrm{c}} {\bm{R}_{\mathrm{i}}^{\mathrm{d}}}^{\mathsf{T}}\right) \end{bmatrix} \nonumber
\end{align}
$\bm{w}_{\mathrm{i}}^{\mathrm{a}}$ is the measured contact wrench at the i-th limb end.
We assume that the limb end is equipped with a 6-axis force sensor.
$\bm{K}_{\mathrm{d}}, \bm{K}_{\mathrm{s}}, \bm{K}_{\mathrm{f}} \in \mathbb{R}^{6 \times 6}$ are diagonal matrices representing damper parameter, spring parameter, and wrench gain, respectively.

For discrete-time control, \eqref{eq:damping-control} is implemented as follows:
\begin{subequations} \label{eq:damping-control-imp}
\begin{align}
  & \Delta \bm{r}_{\mathrm{i,L}}^{\mathrm{c}}[k+1] = \Delta \bm{r}_{\mathrm{i,L}}^{\mathrm{c}}[k] + \Delta t \Delta \bm{\dot{r}}_{\mathrm{i,L}}^{\mathrm{c}}[k] \\
  & \Delta \bm{r}_{\mathrm{i,A}}^{\mathrm{c}}[k+1] = \mathrm{log}\left(\mathrm{exp} \left(\Delta t \Delta \bm{\dot{r}}_{\mathrm{i,A}}^{\mathrm{c}}[k] \times \right) \, \mathrm{exp} \left(\Delta \bm{r}_{\mathrm{i,A}}^{\mathrm{c}}[k] \times \right) \right) \\
  & {\rm where} \ \ \ \ \Delta \bm{\dot{r}}_{\mathrm{i}}^{\mathrm{c}}[k] = - \frac{\bm{K}_{\mathrm{s}}}{\bm{K}_{\mathrm{d}}} \Delta \bm{r}_{\mathrm{i}}^{\mathrm{c}}[k] + \frac{\bm{K}_{\mathrm{f}}}{{\bm{K}_{\mathrm{d}}}} (\bm{w}_{\mathrm{i}}^{\mathrm{a}}[k] - \bm{w}_{\mathrm{i}}^{\mathrm{d}}[k]) \nonumber
\end{align}
\end{subequations}
$\mathrm{exp}(\bm{a}\times) \in \mathbb{R}^{3 \times 3}$ is a function that converts the axis-angle vector $\bm{a}$ to an equivalent rotation matrix.
Since $\bm{K}_{\mathrm{d}}, \bm{K}_{\mathrm{s}}, \bm{K}_{\mathrm{f}}$ are diagonal matrices, their division implies element-wise computation.

%%%%%%%%
%% \subsection{Estimation of Centroidal State}

%% ToDo

%%%%%%%%%%%%%%%%%%%%%%%%%%%%%%%%%%%%%%%%%%%%%%%%%%%%%%%%%%%%%%%%%%%%%%%%%%%%%%%%
\section{Simulation Experiments} \label{sec:exp}

\renewcommand{\arraystretch}{1.25}
\begin{table*}[h]
  \caption{Parameters for preview control in \eqref{eq:pc-objective-linear} \eqref{eq:pc-calc-input-linear}}
  \label{tab:pc-param}
  \vspace{-4mm}
  \begin{center}
    \begin{tabular}{cccccc}
      \hline
      $\bm{Q}_{\mathrm{L}}$ & $\bm{R}_{\mathrm{L}}$ & $\bm{Q}_{\mathrm{A}}$ & $\bm{R}_{\mathrm{A}}$ & $\Delta \tau$ [s] & $N_h$\\
      \hline
      $\mathrm{diag}(\num{2e2}, \num{5e-4})$ & $\mathrm{diag}(\num{1e-8})$ & $\mathrm{diag}(\num{1e2}, \num{5e-3})$ & $\mathrm{diag}(\num{1e-8})$ & 0.005 & $400 (= 2.0 / \Delta \tau)$\\
      \hline
    \end{tabular}\\
    \vspace{2mm}
    \begin{minipage}{1.7\columnwidth}
      \footnotesize{
        $\Delta \tau$ is the discretization period of the horizon.
        $\mathrm{diag}$ denotes the diagonal matrix.
      }
    \end{minipage}
  \end{center}
  \caption{Parameters for damping control in \eqref{eq:damping-control}}
  \label{tab:damping-param}
  \vspace{-4mm}
  \begin{center}
    \begin{tabular}{l||ccc}
      \hline
      & $\bm{K}_{\mathrm{d}}$ & $\bm{K}_{\mathrm{s}}$ & $\bm{K}_{\mathrm{f}}$ \\
      \hline
      Contact phase & $\mathrm{diag}(10000, 10000, 10000, 100, 100, 100)$ & $\mathrm{diag}(0, 0, 0, 0, 0, 2000)$ & $\mathrm{diag}(1, 1, 1, 1, 1, 0)$ \\
      \hline
      Non-contact phase & $\mathrm{diag}(300, 300, 300, 40, 40, 40)$ & $\mathrm{diag}(2250, 2250, 2250, 400, 400, 400)$ & $\mathrm{diag}(0, 0, 0, 0, 0, 0)$ \\
      \hline
    \end{tabular}\\
    \vspace{2mm}
    \begin{minipage}{1.8\columnwidth}
      \footnotesize{
        When $\bm{K}_{\mathrm{f}}$ is zero, the compliance displacement~$\Delta \bm{r}_{\mathrm{i}}^{\mathrm{c}}$ converges linearly to zero from \eqref{eq:damping-control-imp}.
        When a single limb end is in contact, the parameters of the linear component (the first three elements) of the contacting limb end are set to the parameter values for the non-contact phase.
        For the linear component of $\bm{K}_{\mathrm{d}}$ in the contact phase of the hands, different parameter values are used for some motions;
        $1000$ for walking with hands on the wall, and $50000$ for climbing a vertical ladder in \figref{fig:exp-multi-contact}.
        The parameter values are different because they are determined to be as compliant as possible within the range of non-vibration, and the hand contact forces, which differ significantly for each motion, affect the ease of vibration.
      }
    \end{minipage}
  \end{center}
\end{table*}

\renewcommand{\arraystretch}{1.25}
\begin{table}[h]
  \caption{Parameters for centroidal feedback control in \eqref{eq:com-feedback}}
  \label{tab:com-feedback-param}
  \vspace{-4mm}
  \begin{center}
    \begin{tabular}{cc}
      \hline
      $\bm{K}_{\mathrm{P}}$ & $\bm{K}_{\mathrm{D}}$ \\
      \hline
      $\mathrm{diag}(2000, 2000, 2000, 0, 0, 0)$ & $\mathrm{diag}(666, 666, 666, 0, 0, 0)$ \\
      \hline
    \end{tabular}\\
    \vspace{2mm}
    \begin{minipage}{0.85\columnwidth}
      \footnotesize{
        For climbing a vertical ladder in \figref{fig:exp-multi-contact}, the following parameters are used:\\
        $\bm{K}_{\mathrm{P}} = \mathrm{diag}(3000, 3000, 3000, 1000, 1000, 1000)$\\
        $\bm{K}_{\mathrm{D}} = \mathrm{diag}(1000, 1000, 1000, 333, 333, 333)$
      }
    \end{minipage}
  \end{center}
\end{table}

%%%%%%%%
\subsection{Robot Controller Implementation}

\subsubsection{Software Framework}

The proposed control system is implemented in {\tt C++} within a real-time robot control framework {\tt mc\_rtc}~\cite{mc_rtc:github2021}.
QLD~\cite{QLD:scilab2021} is used as the QP solver for wrench distribution.
For the preview control, we only need to implement \eqref{eq:pc-calc-input-linear}, and no external library is required.

Kinematics commands, such as the CoM position and base link orientation $\bm{r}^{\mathrm{d}}$ and the limb end position and orientation $\bm{p}_{\mathrm{i}}^{\mathrm{c}}, \bm{R}_{\mathrm{i}}^{\mathrm{c}}$, are passed to the acceleration-based whole-body inverse kinematics (IK) calculation.
A low stiffness task for nominal joint position is also imposed in IK, so that the joints of the limb without contact approach to the nominal position.
The calculated joint angles are commanded to the low-level joint position PD controller.
As sensor measurements, the proposed control system uses the joint angles from the joint encoders, the contact wrench from the 6-axis force sensors mounted on the hands and feet, and the link orientation from the inertial measurement unit (IMU) sensor mounted on the trunk link.

\subsubsection{Controller Parameters}

Tables~\ref{tab:pc-param}, \ref{tab:damping-param}, and \ref{tab:com-feedback-param} show the control parameters.
For these parameters, the same values are used for bipedal walking and various multi-contact motions presented in this section, except for $\bm{K}_{\mathrm{d}}$ of hand damping control shown in Table~\ref{tab:damping-param}.

\subsubsection{Simulation Environments}

We verified various multi-contact motions of the humanoid robot HRP-5P~\cite{HRP5P:Kaneko:RAL2019}, which is developed by the authors' group, on the dynamics simulator Choreonoid~\cite{Choreonoid:Nakaoka:SII2012}.
The control processes shown in \figref{fig:system} are executed every 2 ms on a single thread with an Intel Core i7-9750H CPU (2.60 GHz) and 32 GB 2667 MHz RAM.
We also confirmed that some of the motions can be executed on the dynamics simulator MuJoCo~\cite{MuJoCo:Todorov:IROS2012} and on other humanoid robots in the HRP series, in order to verify the robustness of the simulation engines and robot models.
See the supplemental video for all the experiments.

%%%%%%%%
\subsection{Bipedal Walking}

\begin{figure}[tpb]
  \begin{center}
    \includegraphics[width=\columnwidth, clip, bb=150 10 1755 410]{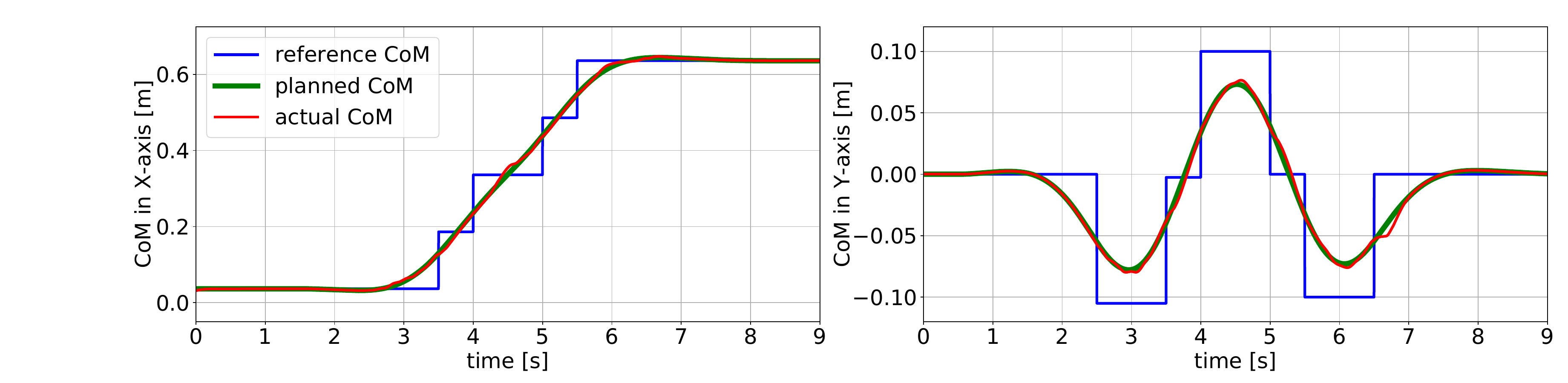}\\
    \footnotesize{(A) CoM}\\
    \vspace{2mm}
    \includegraphics[width=\columnwidth, clip, bb=150 10 1755 410]{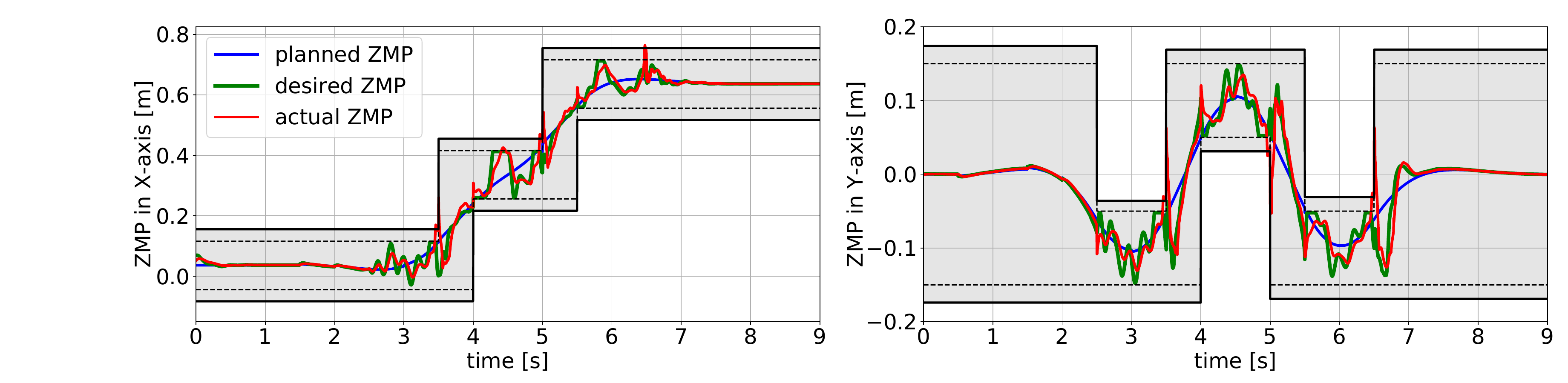}\\
    \footnotesize{(B) ZMP}
    \caption{Results of bipedal walking.
      \newline
      \footnotesize{
        (A) The reference, planned, and actual CoMs correspond to $\bm{r}^{\mathrm{ref}}$, $\bm{r}^{\mathrm{p}}$, and $\bm{r}^{\mathrm{a}}$ in \figref{fig:system}, respectively.
        (B) The planned, desired, and actual ZMPs are calculated from $\bm{\bar{w}}^{\mathrm{p}^\prime}$, $\bm{w}^{\mathrm{d}}_{\mathrm{i}}$, and $\bm{w}_{\mathrm{i}}^{\mathrm{a}}$ in \figref{fig:system}, respectively.
        The gray shaded region indicates the support region.
        The wrench distribution uses contact polygon vertices with inner margins, and the black dashed lines indicate their boundaries.
    }}
    \label{fig:exp-walk-graph}
  \end{center}
\end{figure}

First, the robot performed bipedal walking on uneven floors, stairs, and ramps.
\figref{fig:exp-walk-graph} shows the graphs of CoM and ZMP when walking straight on a flat ground with a stride length of 300~mm.
The ZMP is not used in the control but is calculated from the resultant wrench for verification only.
The reference CoM trajectory is determined by a simple rule; the horizontal position is the center of the supporting foot in the single support phase, the center of both feet in the double support phase, and the vertical position is constant.
The planned CoM trajectory shown in \figref{fig:exp-walk-graph}~(A) is similar to the CoM trajectories planned by the conventional bipedal walking methods~\cite{PreviewControl:Kajita:ICRA2003}, although ZMP tracking is not explicitly considered in the proposed method.
\figref{fig:exp-walk-graph}~(B) shows that the planned ZMP falls within the foot support region, and the ZMP modified by the feedback is tracked by the simulated robot.

%% \begin{figure}[tpb]
%%   \begin{center}
%%     \includegraphics[width=0.48\columnwidth]{figs/walk-uneven.png}
%%     \hspace{1mm}
%%     \includegraphics[width=0.48\columnwidth]{figs/stair-slope.png}\\
%%     \begin{minipage}{0.48\columnwidth}
%%       \begin{center} \footnotesize (A) Uneven floor \end{center}
%%     \end{minipage}
%%     \hspace{1mm}
%%     \begin{minipage}{0.48\columnwidth}
%%       \begin{center} \footnotesize (B) Stairs and a ramp \end{center}
%%     \end{minipage}
%%     \caption{Simulation of bipedal walking.
%%       \newline
%%       \footnotesize{
%%         (A) 10~mm uneven floor. (B) Four steps of 50~mm height and a 10~degree ramp.
%%     }}
%%     \label{fig:exp-walk}
%%   \end{center}
%% \end{figure}

\begin{figure*}[tpb]
  \begin{center}
    \includegraphics[width=0.385\columnwidth]{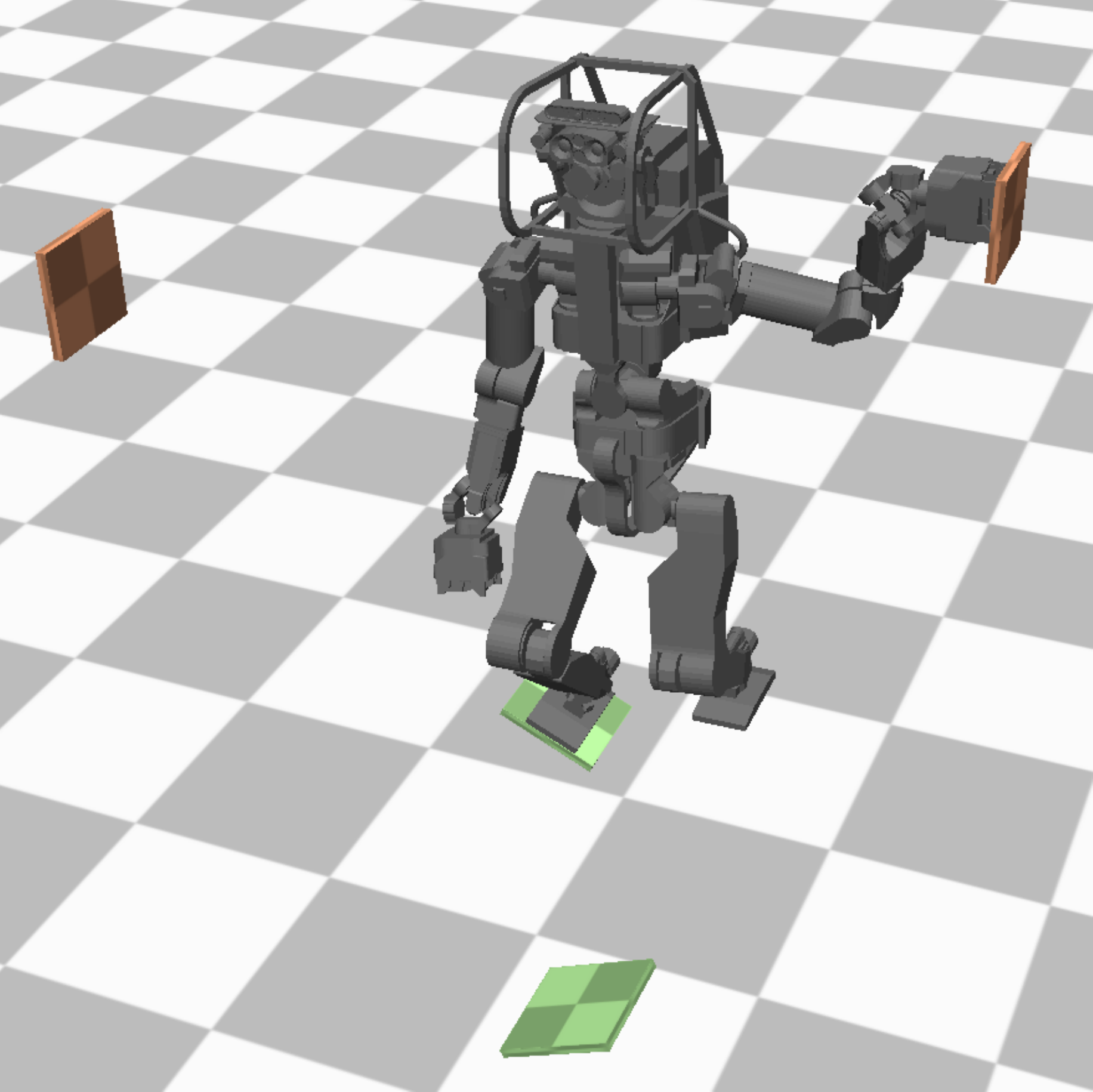}
    \hspace{0.1mm}
    \includegraphics[width=0.385\columnwidth]{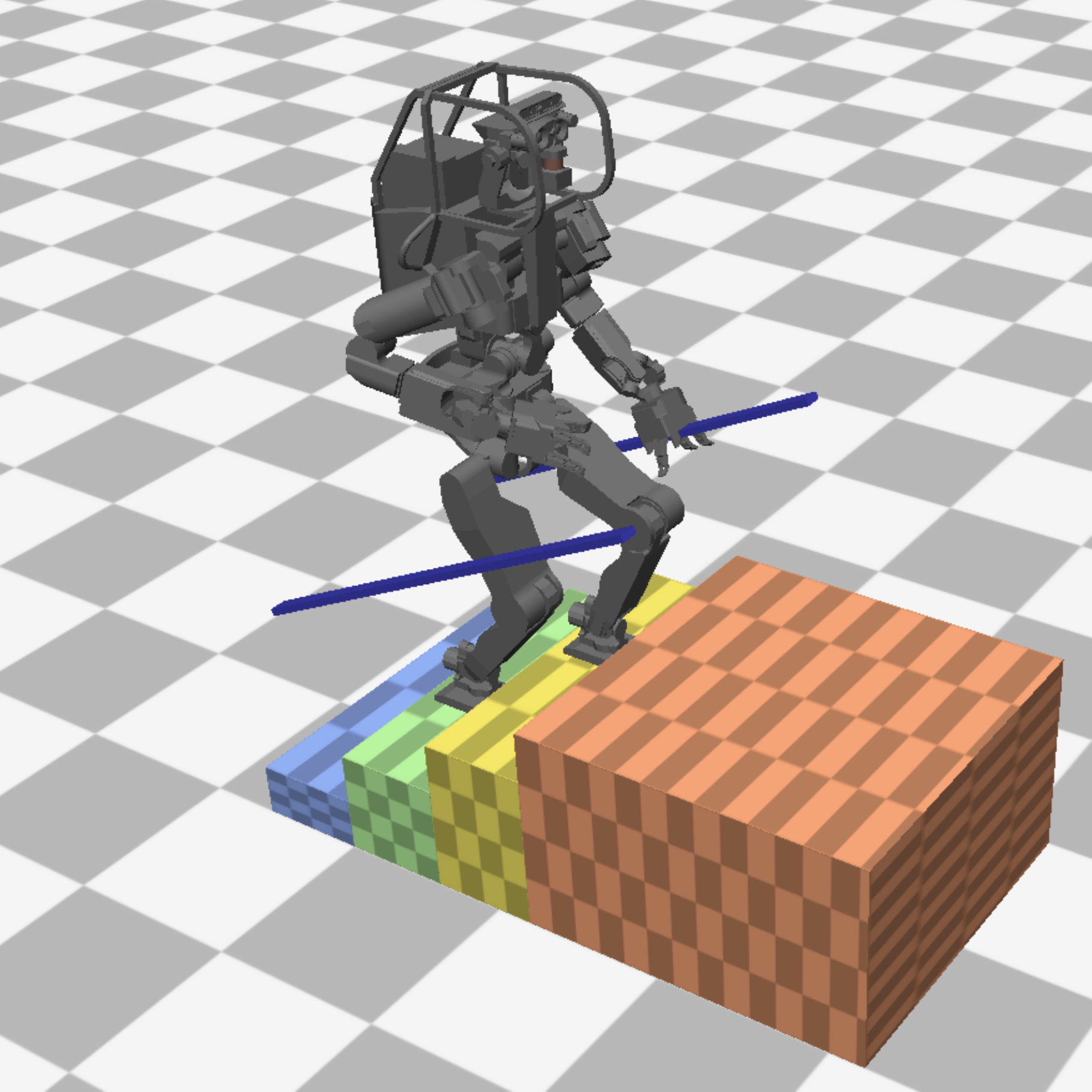}
    \hspace{0.1mm}
    \includegraphics[width=0.385\columnwidth]{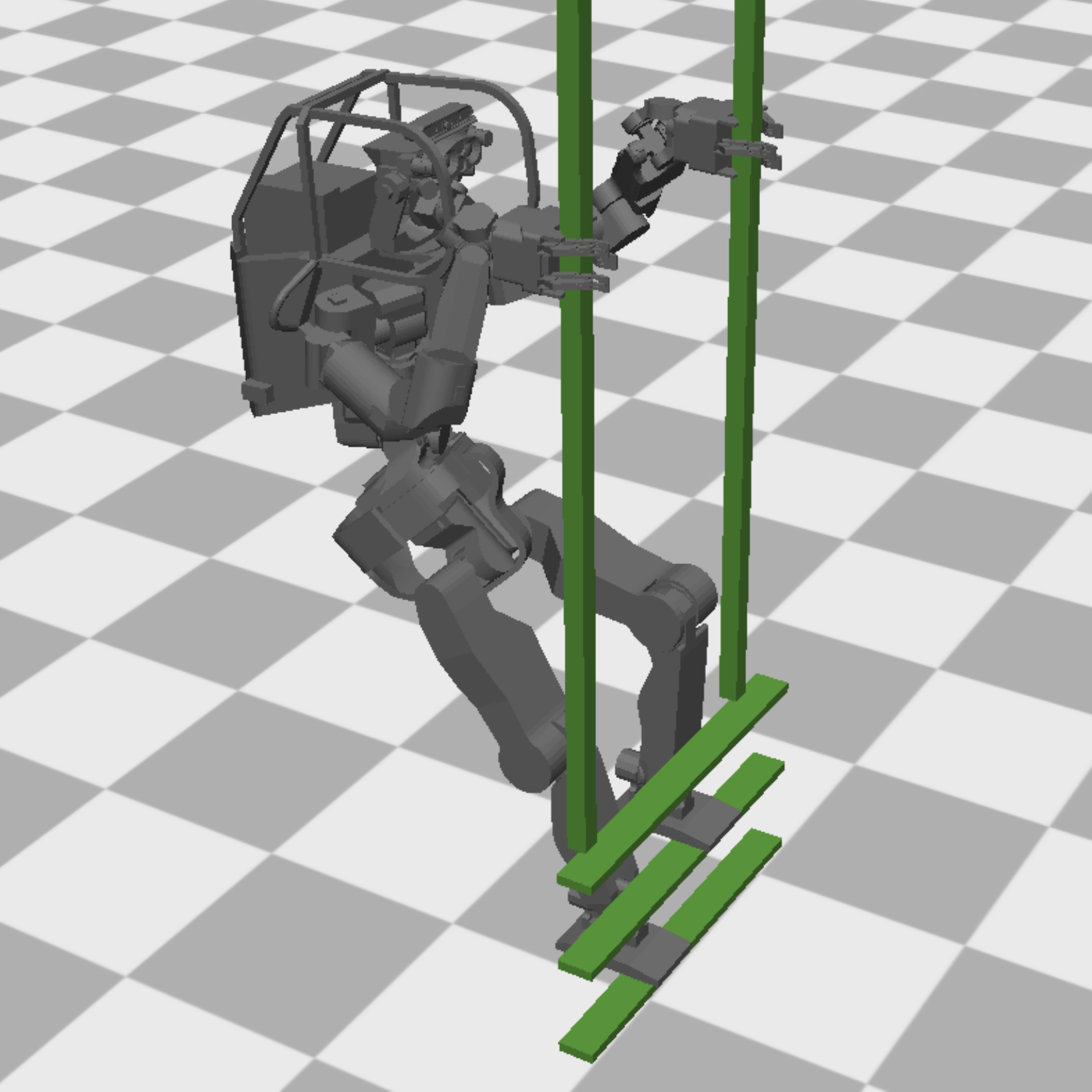}
    \hspace{0.1mm}
    \includegraphics[width=0.385\columnwidth]{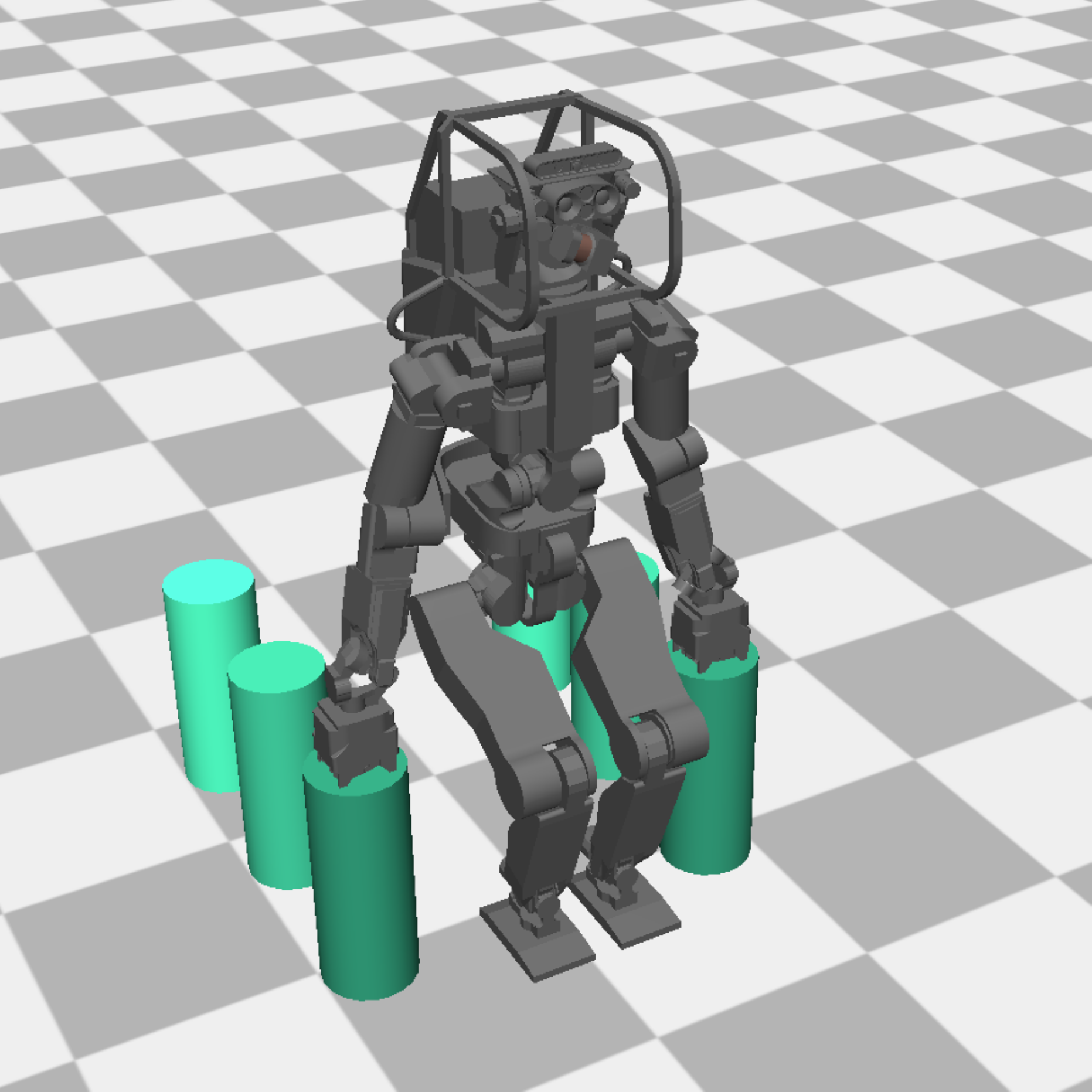}
    \hspace{0.1mm}
    \includegraphics[width=0.385\columnwidth]{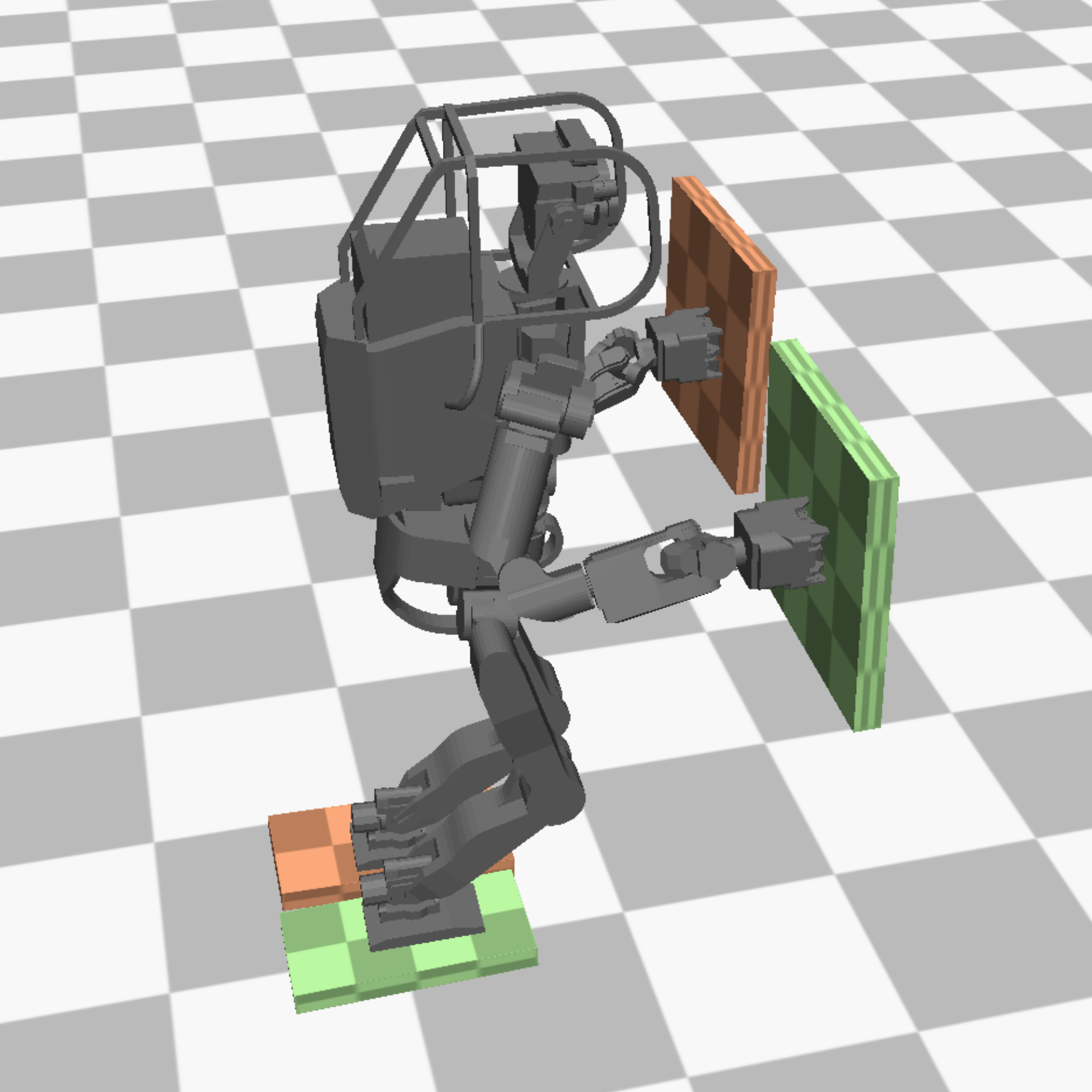}\\
    \vspace{1mm}
    \begin{minipage}{0.385\columnwidth}
      \begin{center} \footnotesize (A) Walking with hands \\on the wall \end{center}
    \end{minipage}
    \hspace{0.1mm}
    \begin{minipage}{0.385\columnwidth}
      \begin{center} \footnotesize (B) Climbing handrail stairs \end{center}
    \end{minipage}
    \hspace{0.1mm}
    \begin{minipage}{0.385\columnwidth}
      \begin{center} \footnotesize (C) Climbing a vertical ladder \end{center}
    \end{minipage}
    \hspace{0.1mm}
    \begin{minipage}{0.385\columnwidth}
      \begin{center} \footnotesize (D) Moving with both hands supporting the body \end{center}
    \end{minipage}
    \hspace{0.1mm}
    \begin{minipage}{0.385\columnwidth}
      \begin{center} \footnotesize (E) Contact with moving floors and walls \end{center}
    \end{minipage}
    \caption{Simulation of multi-contact motion.
      \newline
      \footnotesize{
        The friction coefficient is set to 0.6 for all environments.
        (A) Stepping on scaffold boards with 25~degree incline while keeping the hand on a vertical wall.
        (B) Climbing four steps of 150~mm height with handrails.
        (C) Climbing a vertical ladder with 200~mm steps. The depth of the ladder rungs is 75~mm, which is less than half the depth of the robot's sole.
        (D) With both hands on the support surfaces at the height of 500~mm, moving forward by swinging both feet simultaneously.
        (E) Balancing on the floors and walls, which move periodically in translation and rotation with an amplitude of 20~mm and 2~degrees.
        Damping control allows feet and hands to adapt to floor and wall movements.
    }}
    \label{fig:exp-multi-contact}
  \end{center}
\end{figure*}

%%%%%%%%
\subsection{Multi-contact Motion}

\begin{figure}[tpb]
  \begin{center}
    \includegraphics[width=\columnwidth, clip, bb=60 12 1700 410]{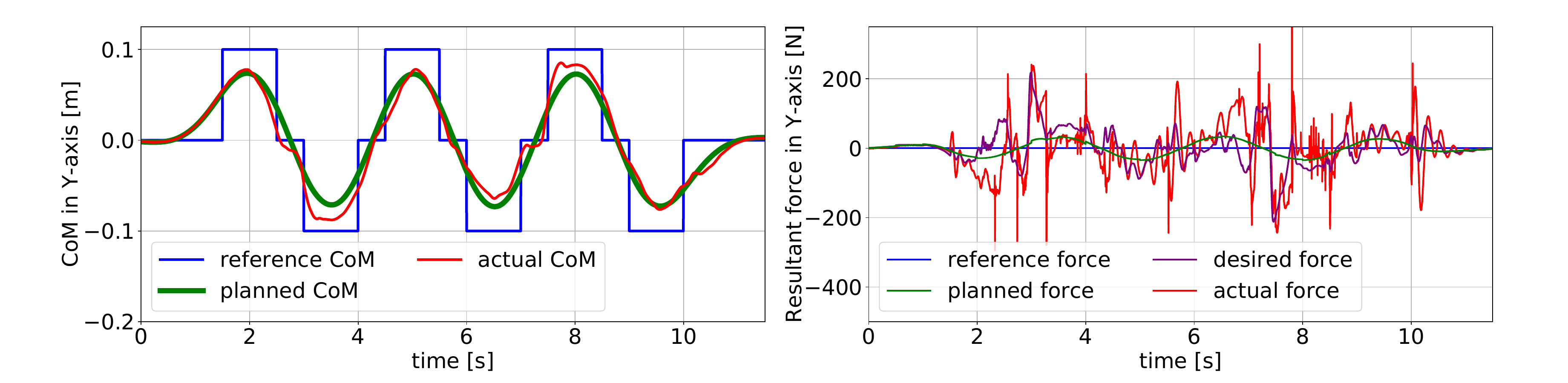}\\
    \begin{minipage}{0.45\columnwidth}
      \begin{center} \footnotesize (A) CoM \end{center}
    \end{minipage}
    \hspace{1mm}
    \begin{minipage}{0.45\columnwidth}
      \begin{center} \footnotesize (B) Resultant wrench \end{center}
    \end{minipage}\\
    \vspace{2mm}
    \includegraphics[width=\columnwidth, clip, bb=60 12 1700 410]{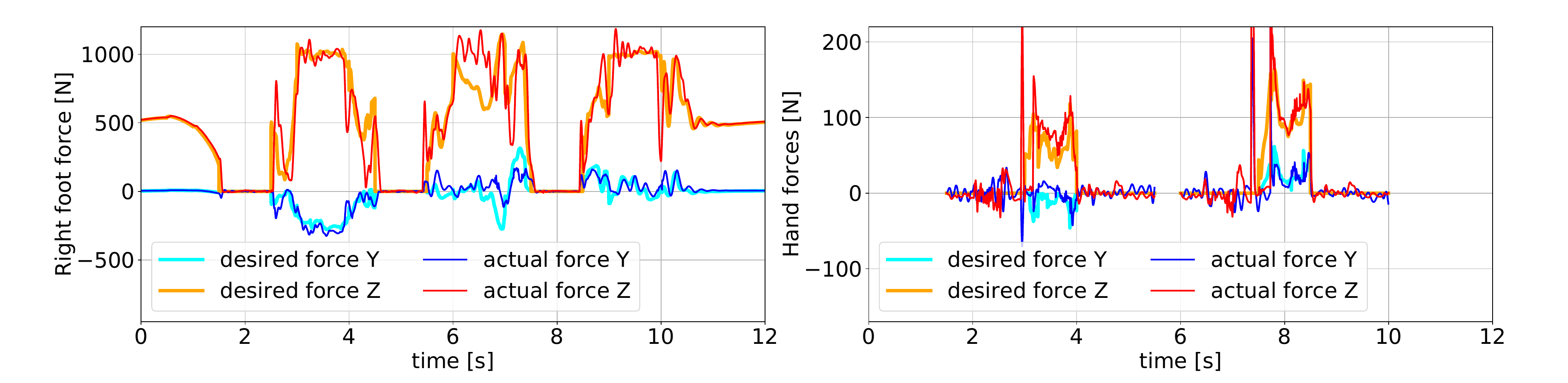}\\
    \begin{minipage}{0.45\columnwidth}
      \begin{center} \footnotesize (C) Foot contact forces \end{center}
    \end{minipage}
    \hspace{1mm}
    \begin{minipage}{0.45\columnwidth}
      \begin{center} \footnotesize (D) Hand contact forces \end{center}
    \end{minipage}
    \caption{Results of walking with hands on the wall in \figref{fig:exp-multi-contact}~(A).
      \newline
      \footnotesize{
        The CoM and resultant wrench are represented in the world coordinates, and the contact forces are represented in the coordinates at each limb end (see \figref{fig:wrench-distrib}).
        (B) The reference, planned, desired, and actual resultant forces are calculated from $\bar{f}^{\mathrm{ref}}_*$, $\bm{\bar{w}}^{\mathrm{p}^\prime}$, $\bm{w}^{\mathrm{d}}_{\mathrm{i}}$, and $\bm{w}^{\mathrm{a}}_{\mathrm{i}}$, respectively.
        (C) The desired and actual forces correspond to $\bm{w}^{\mathrm{d}}_{\mathrm{i}}$ and $\bm{w}^{\mathrm{a}}_{\mathrm{i}}$, respectively.
        (D) Contact forces of the left hand (1-6~s) and right hand (6-10~s).
    }}
    \label{fig:exp-walk-with-hand}
  \end{center}
\end{figure}

\begin{figure}[tpb]
  \begin{center}
    \vspace{2mm}
    \includegraphics[width=\columnwidth, clip, bb=60 10 1748 410]{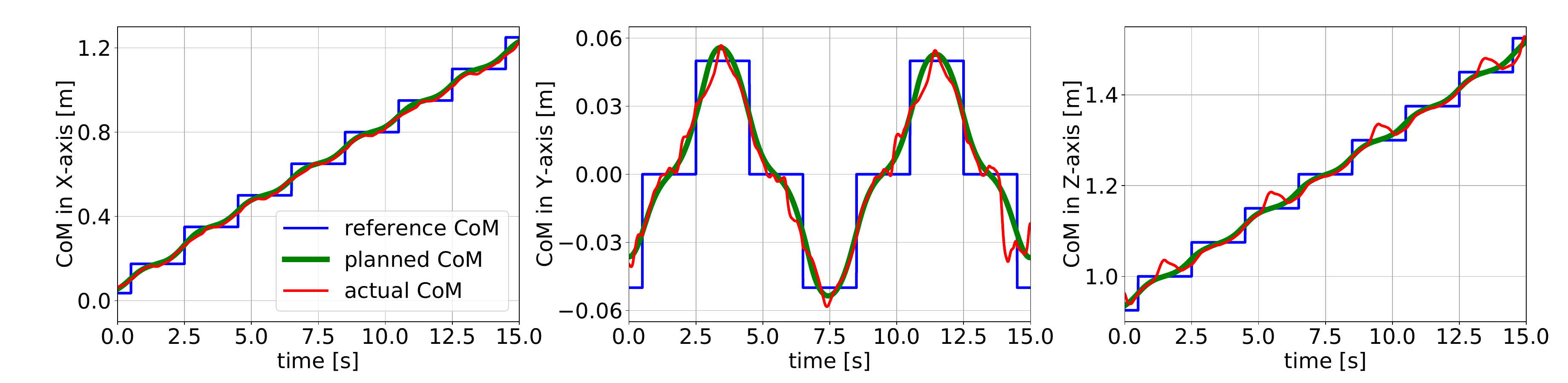}\\
    \vspace{-0.5mm}
    \footnotesize{(A) CoM}\\
    \vspace{1.5mm}
    \includegraphics[width=\columnwidth, clip, bb=60 10 1687 410]{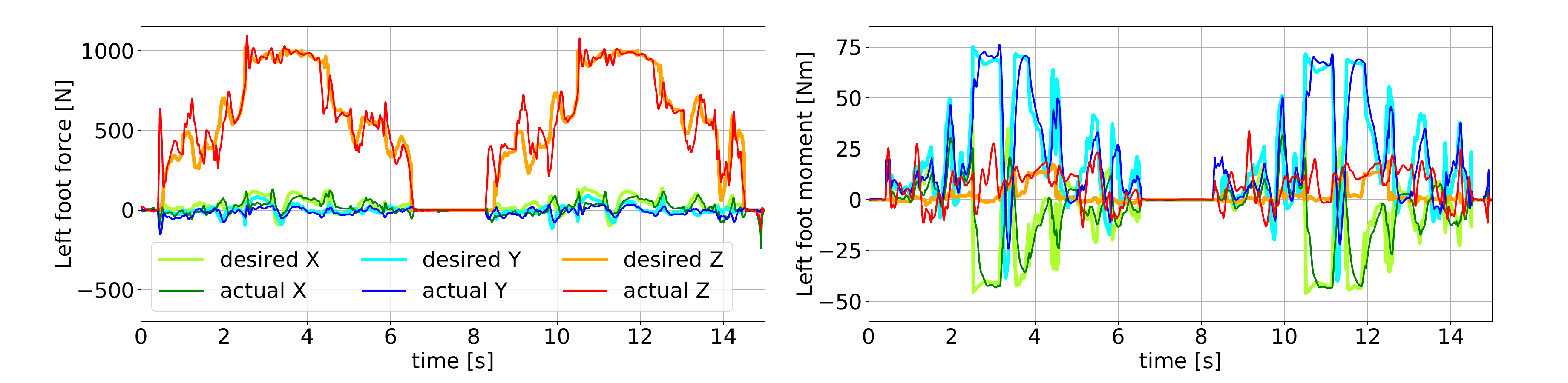}\\
    \vspace{-0.5mm}
    \footnotesize{(B) Foot contact wrenches}\\
    \vspace{1.5mm}
    \includegraphics[width=\columnwidth, clip, bb=60 10 1687 410]{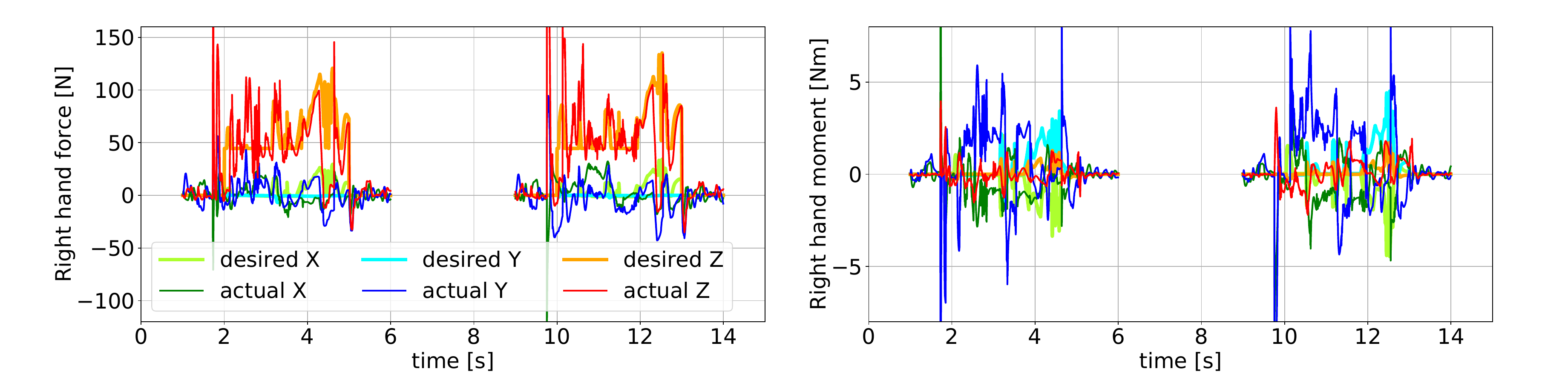}\\
    \vspace{-0.5mm}
    \footnotesize{(C) Hand contact wrenches}
    \caption{Results of climbing handrail stairs in \figref{fig:exp-multi-contact}~(B).
      \newline
      \footnotesize{
        Vertical forces of about 50-100~N are applied to the hands.
        The error in the Z position of the actual CoM is because the result of the whole-body inverse kinematics cannot follow the desired CoM due to the limitation of kinematics reachability.
    }}
    \label{fig:exp-handrail-stairs}
%%   \end{center}
%% \end{figure}
%% \begin{figure}[tpb]
%%   \begin{center}
    \vspace{4mm}
    \includegraphics[width=\columnwidth, clip, bb=25 10 1748 410]{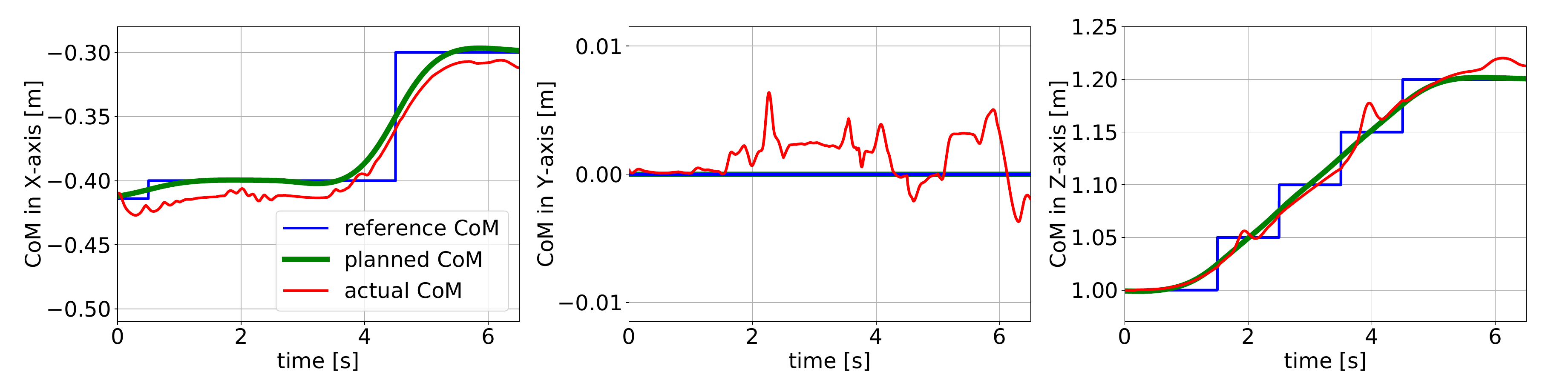}\\
    \vspace{-0.5mm}
    \footnotesize{(A) CoM}\\
    \vspace{1.5mm}
    \includegraphics[width=\columnwidth, clip, bb=40 10 1687 410]{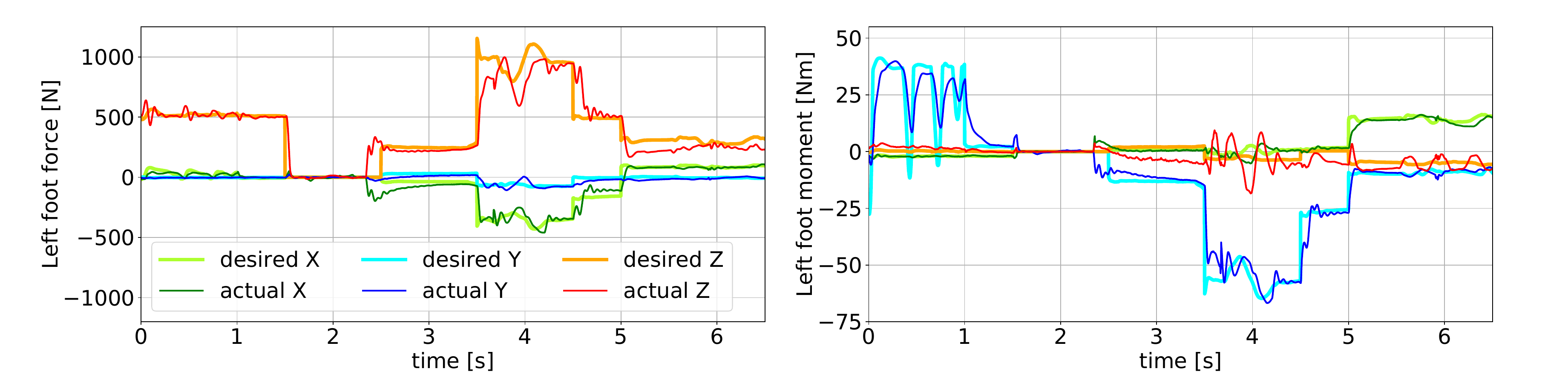}\\
    \vspace{-0.5mm}
    \footnotesize{(B) Foot contact wrenches}\\
    \vspace{1.5mm}
    \includegraphics[width=\columnwidth, clip, bb=40 10 1687 410]{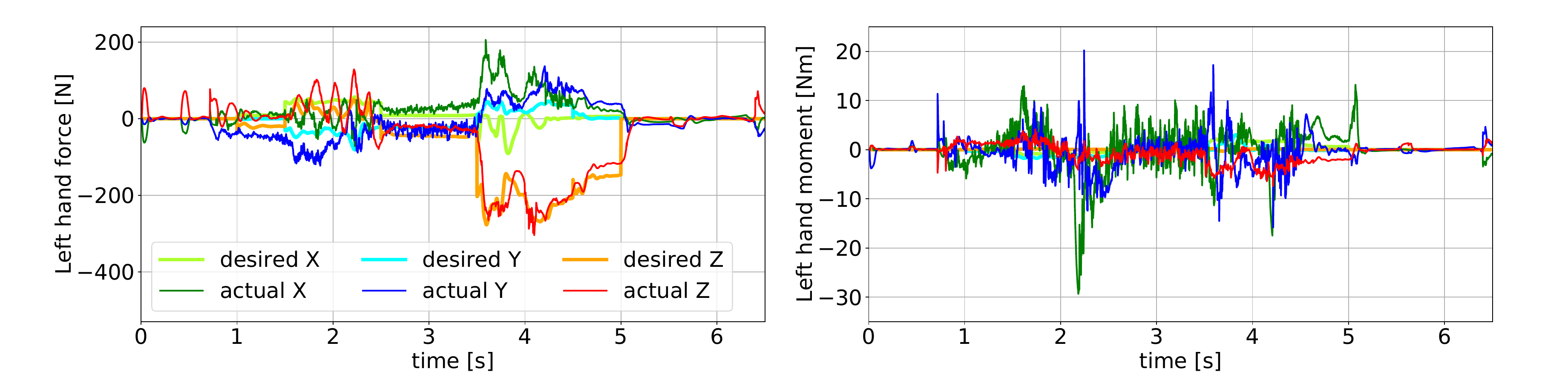}\\
    \vspace{-0.5mm}
    \footnotesize{(C) Hand contact wrenches}
    \caption{Results of climbing a vertical ladder in \figref{fig:exp-multi-contact}~(C).
      \newline
      \footnotesize{
        Pulling forces of about 200~N are applied to the hands.
        Since the foot is in contact with the ladder rung only at the toes, the foot exerts a large contact moment around the Y-axis.
    }}
    \label{fig:exp-vertical-ladder}
%%   \end{center}
%% \end{figure}
%% \begin{figure}[tpb]
%%   \begin{center}
    \vspace{4mm}
    \includegraphics[width=0.48\columnwidth]{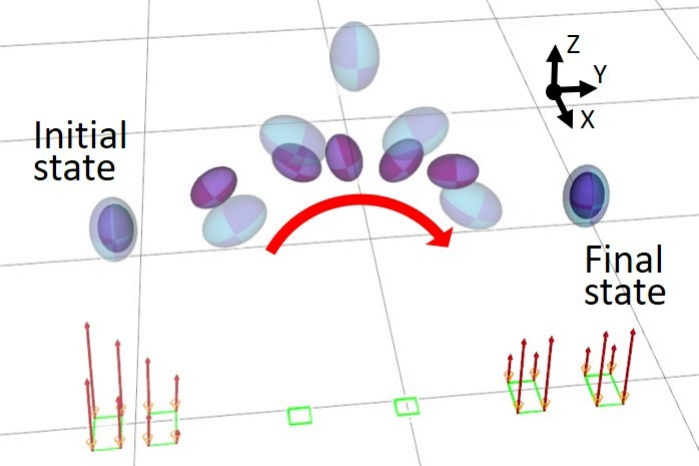}
    \hspace{1mm}
    \includegraphics[width=0.48\columnwidth, clip, bb=58 10 698 425]{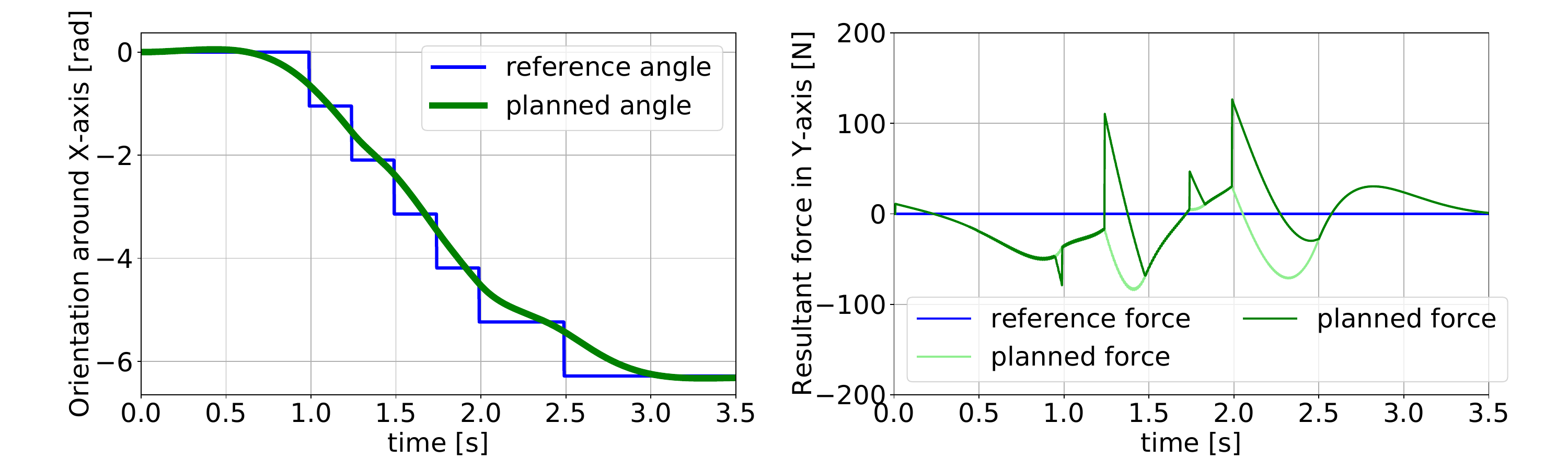}\\
    \vspace{-0.5mm}
    \begin{minipage}{0.48\columnwidth}
      \begin{center} \footnotesize (A) Outline of centroidal trajectory \end{center}
    \end{minipage}
    \begin{minipage}{0.48\columnwidth}
      \begin{center} \footnotesize (B) Base link orientation \end{center}
    \end{minipage}
    \caption{Centroidal trajectory generation of cartwheel motion.
      \newline
      \footnotesize{
        (A) The translucent markers represent the reference, and the non-translucent markers represent the planned centroidal states, respectively.
        Green rectangles represent the contact positions of the hands and feet.
    }}
    \label{fig:exp-cartwheel}
  \end{center}
\end{figure}

We applied the proposed control methods to the five types of multi-contact motions shown in \figref{fig:exp-multi-contact}.
As shown in the supplemental video, various motions with different contact transition orders, hand forces, and contact forms (i.e., unilateral contact or grasping contact) are stably performed.
Figs.~\ref{fig:exp-walk-with-hand}, \ref{fig:exp-handrail-stairs}, and \ref{fig:exp-vertical-ladder} show the graphs of the motion results of \figref{fig:exp-multi-contact} (A), (B), and (C), respectively.
It can be observed that the desired contact wrenches distributed to the limb ends are tracked by the damping control, and the simulated robot tracks the planned CoM trajectory.
%% The hands are subjected to forces of about 50-100~N for walking with hands on the wall and climbing handrail stairs, and about 200~N for climbing a vertical ladder.
The reference CoM trajectories are represented by piecewise-constant functions synchronized with the contact switching as Figs.~\ref{fig:exp-walk-with-hand}, \ref{fig:exp-handrail-stairs}, and \ref{fig:exp-vertical-ladder} show.
Specifically, in the motions of \figref{fig:exp-multi-contact}~(A) and (B), the reference CoM position is determined by the same rule as in bipedal walking described in the previous section.
In the motion of \figref{fig:exp-multi-contact}~(B), the CoM lateral position in the single support phase is offset inward by 50~mm so that the hands support part of the robot's weight.
In the motion of \figref{fig:exp-multi-contact}~(C), the reference CoM position is set 0.4~m behind the center of the ladder, but offset forward by 0.1~m to reduce the pitch moment during the hand contact transitions.
In all motions, the CoM vertical position is determined by adding a constant offset to the mean of the foot contact positions.
The reference base link orientations are all set to zero.
%% The predefined reference CoM trajectories shown in Figs.~\ref{fig:exp-walk-with-hand}, \ref{fig:exp-handrail-stairs}, and \ref{fig:exp-vertical-ladder} are piecewise-constant according to the contact transitions and do not contain an arbitrary tuning, so the proposed methods, including the determination of the reference trajectories, are generic.

%%%%%%%%
\subsection{Motion to Change the Base Link Orientation}

The proposed CoM trajectory generation was applied to a cartwheel motion with a large change in base link orientation, as shown in \figref{fig:exp-cartwheel}.
From a rough reference centroidal trajectory and given contact positions, a dynamically feasible centroidal trajectory and contact wrenches are planned such that the base link makes one revolution while moving laterally.
Since mapping this centroidal motion to the motion in the joint configuration space is difficult for a simple whole-body inverse kinematics calculation, this paper only verifies the centroidal trajectory generation, and its execution on the robot is out of scope.

%%%%%%%%
\subsection{Computation Time}

Table~\ref{tab:comp-time} shows the computation time of the proposed control method for the multi-contact motions in \figref{fig:exp-multi-contact}.
It takes 0.1-0.4~ms for one control update, depending on the number of robot contacts.
The MPC-based methods take 20~ms for a 3~s horizon (20 sample points)~\cite{MultiContact:Audren:IROS2014} and 1~ms for a 0.5~s horizon (10 sample points)~\cite{HumanoidMPC:Henze:IROS2014} for each control update, whereas our method takes 0.4~ms for a 2~s horizon (400 sample points).
This shows that our method reduces the computation time to about half or less while increasing the number of sample points in the horizon by about 20 times.

\renewcommand{\arraystretch}{1.0}
\begin{table}[t]
  \caption{Computation time [$\mathrm{\mu}$s]}
  \label{tab:comp-time}
  \vspace{-4mm}
  \begin{center}
    \begin{tabular}{l||l|l|l}
      \hline
      & (A) & (B) & (C) \\
      \hline
      Total control & 124 & 186 & 409 \\
      \hline
      \hspace{1mm} Calculate reference & 22 (18 \%) & 22 (12 \%) & 22 (5 \%) \\
      \hline
      \hspace{1mm} Preview control & 9 \ (7 \%) & 10 \ (5 \%) & 9 \ (2 \%) \\
      \hline
      \hspace{1mm} Project planned wrench & 29 (24 \%) & 56 (30 \%) & 165 (40 \%) \\
      \hline
      \hspace{1mm} Distribute desired wrench & 28 (22 \%) & 56 (30 \%) & 162 (40 \%) \\
      \hline
    \end{tabular}\\
    \vspace{2mm}
    \begin{minipage}{0.95\columnwidth}
      \footnotesize{
        The computation time per control cycle for the motions (A), (B), and (C) in \figref{fig:exp-multi-contact} is shown.
        In the computational time measurements, only the proposed centroidal control is included, not the whole-body inverse kinematics.
        The higher the number of robot contacts, the more time is required for wrench projection and distribution, but the time for the other processes remains almost unchanged.
      }
    \end{minipage}
  \end{center}
\end{table}

%%%%%%%%
\subsection{Validation of Rotational Motion Approximation} \label{sec:val-rot-motion}

\subsubsection{Approximation of Angular Momentum}

To validate the approximation of the centroidal angular momentum, we calculated the errors between $\bm{I}_{\mathrm{all}} \, \bm{\dot{q}}$, $\bm{I}_{\mathrm{base}} \, \bm{\omega}$, and $\bm{I} \bm{\omega}$ in \eqref{eq:angular-momentum-approx} for the motions of bipedal walking and climbing handrail stairs.
The mean error was 0.044~kg\,m\textsuperscript{2}/s for $\|\bm{I}_{\mathrm{base}} \, \bm{\omega} - \bm{I} \bm{\omega}\|$ and 3.1~kg\,m\textsuperscript{2}/s for $\|\bm{I}_{\mathrm{all}} \, \bm{\dot{q}} - \bm{I} \bm{\omega}\|$.
Since the robot did not fall over in the simulation, these errors were not enough to break the motion feasibility.
Furthermore, by adding the angular momentum task~\cite{CentroidalDynamics:Orin:AuRo2013} to the whole-body IK, the mean error was reduced to 0.011~kg\,m\textsuperscript{2}/s for $\|\bm{I}_{\mathrm{all}} \, \bm{\dot{q}} - \bm{I} \bm{\omega}\|$.

\subsubsection{Approximation of Angular Velocity}

In \eqref{eq:pc-state-eq-angular}, the angular velocity of the base link is approximated to the time derivative of the Euler angle.
To validate this approximation, we calculated the error between $\bm{I} \bm{\omega}$ and $\bm{I} \bm{\dot{\alpha}}$.
The mean error for the same two motions as in the previous section was 0.0051~kg\,m\textsuperscript{2}/s.
The small error was due to the small inclination of the base link during the motions.
Additionally, the error was small even in the cartwheel motion because $\bm{K}_{\mathrm{Euler}}$ was not affected by rotation around the X-axis in the ZYX Euler angle used in our implementation.

%%%%%%%%
\subsection{Comparison with Constrained MPC}

The proposed centroidal trajectory generation consists of unconstrained preview control and post-processing wrench projection.
This greatly reduces computational cost, but the generated trajectory may be far from the global optimum.
We therefore compared the centroidal trajectories generated by the proposed method and the linear MPC-based method~\cite{MultiContact:Audren:IROS2014} for the motions of climbing handrail stairs and climbing a vertical ladder.
The mean projection errors of the planned resultant wrench onto the contact constraint manifold (i.e., $\| \bm{\bar{w}}^{\mathrm{p}} - \bm{\bar{w}}^{\mathrm{p}^\prime} \|$) were 2.2~N and 4.4~Nm for climbing handrail stairs and zero for climbing a vertical ladder in the proposed method.
These projection errors were zero in the MPC-based method because the trajectories consistent with the constraints were planned.
The mean projection errors of the desired resultant wrench $\bm{\bar{w}}^{\mathrm{d}}$ were as follows; for climbing handrail stairs, 3.7~N and 4.7~Nm in the proposed method and 0.42~N and 0.52~Nm in the MPC-based method; and for climbing a vertical ladder, 9.0~N and 8.7~Nm in the proposed method and 9.3~N and 9.1~Nm in the MPC-based method.
No significant differences were found in the scale of feedback wrench~\eqref{eq:com-feedback}, CoM acceleration, and angular momentum.
Based on the above, we conclude that there is no significant difference between the two methods for motions incorporating feedback.
Note that the MPC-based method took more than twice as long as the motion duration.

%%%%%%%%%%%%%%%%%%%%%%%%%%%%%%%%%%%%%%%%%%%%%%%%%%%%%%%%%%%%%%%%%%%%%%%%%%%%%%%%
\section{Conclusion}

In this paper, we proposed a centroidal online trajectory generation and stabilization control for humanoid dynamic multi-contact motion.
The controller combines preview control, which does not explicitly consider constraints, with wrench distribution to satisfy contact constraints, and rapidly generates feasible robot motions.
Simulation experiments showed that various motions such as bipedal walking and climbing handrail stairs and vertical ladders can be achieved stably by a humanoid robot.
The proposed method can update the robot centroidal state in a shorter period for a more finely sampled control horizon than previous methods.

Future works include the online trajectory generation in the joint space in the form of a receding horizon to enhance the inverse kinematics calculation.
This will enable the robot to perform kinematically and dynamically challenging motions such as cartwheels, for which the proposed method generated the centroidal trajectory.
Another future direction is to improve the robustness and reliability using a hierarchical scheme that executes the MPC at low periods and the proposed preview control at high periods.

%% \addtolength{\textheight}{-12cm}   % This command serves to balance the column lengths
%%                                   % on the last page of the document manually. It shortens
%%                                   % the textheight of the last page by a suitable amount.
%%                                   % This command does not take effect until the next page
%%                                   % so it should come on the page before the last. Make
%%                                   % sure that you do not shorten the textheight too much.

%%%%%%%%%%%%%%%%%%%%%%%%%%%%%%%%%%%%%%%%%%%%%%%%%%%%%%%%%%%%%%%%%%%%%%%%%%%%%%%%

%%%%%%%%%%%%%%%%%%%%%%%%%%%%%%%%%%%%%%%%%%%%%%%%%%%%%%%%%%%%%%%%%%%%%%%%%%%%%%%%
\section*{APPENDIX}

\subsection{Relationship with DCM-based Bipedal Control} \label{sec:biped-control}

The proportional control of DCM~\cite{DcmWalk:Englsberger:TRO2015,StairClimb:Caron:ICRA2019} for bipedal walking with constant CoM height and coplanar feet is expressed by the following equation:
\begin{align}
  \bm{z}^{\mathrm{c}} = \bm{z}^{\mathrm{d}} + \bm{K}_{\mathrm{\xi}} (\bm{\xi}^{\mathrm{a}} - \bm{\xi}^{\mathrm{d}})
\end{align}
$\bm{z} \in \mathbb{R}^{2}$ is the ZMP, and $\bm{\xi} \in \mathbb{R}^{2}$ is the DCM.
$\bm{K}_{\mathrm{\xi}} \in \mathbb{R}^{2 \times 2}$ is the diagonal matrix of the feedback gain.
The superscripts c, d, and a stand for command, desired, and actual, respectively.
The command ZMP is converted to the command CoM acceleration and then to the command resultant force as follows:
\begin{subequations}
\begin{align}
  \bm{\ddot{c}}^{\mathrm{c}} &= \omega^2 (\bm{c}^{\mathrm{d}} - \bm{z}^{\mathrm{c}}) \\
  &= \bm{\ddot{c}}^{\mathrm{d}} + \omega^2 \bm{K}_{\mathrm{\xi}} (\bm{\xi}^{\mathrm{d}} - \bm{\xi}^{\mathrm{a}}) \\
  \bm{f}^{\mathrm{c}} &= m \bm{\ddot{c}}^{\mathrm{c}} \\
  &= \bm{f}^{\mathrm{d}}
  + m \omega^2 \bm{K}_{\mathrm{\xi}} (\bm{c}^{\mathrm{d}} - \bm{c}^{\mathrm{a}})
  + m \omega \bm{K}_{\mathrm{\xi}} (\bm{\dot{c}}^{\mathrm{d}} - \bm{\dot{c}}^{\mathrm{a}}) \label{eq:com-feedback-biped}
\end{align}
\end{subequations}
$\omega \in \mathbb{R}$ is the LIPM frequency.
The LIPM dynamics ($\bm{\ddot{c}} = \omega^2 (\bm{c} - \bm{z})$) and the DCM definition ($\bm{\xi} \!=\! \bm{c} \!+\! \frac{1}{\omega} \bm{\dot{c}}$) are used in the equation transformations.
\eqref{eq:com-feedback-biped} has the same form as the linear part of \eqref{eq:com-feedback-2}.
The correspondence of the coefficients is as follows:
\begin{align}
  \bm{K}_{\mathrm{P_L}} = m \omega^2 \bm{K}_{\mathrm{\xi}},\ \ \ \ \ \
  \bm{K}_{\mathrm{D_L}} = m \omega \bm{K}_{\mathrm{\xi}}
\end{align}

This helps to estimate the feedback gains of \eqref{eq:com-feedback} in the proposed method.
For the HRP-5P, $m = 105, \omega^2 = 9.8 / 0.95 = 10.3, \bm{K}_{\mathrm{\xi}} = \mathrm{diag}(2, 2)$,
and therefore $\bm{K}_{\mathrm{P_L}} = \mathrm{diag}(2163, 2163), \bm{K}_{\mathrm{D_L}} = \mathrm{diag}(674, 674)$,
which roughly matches the values in Table~\ref{tab:com-feedback-param}.

%% %%%%%%%%%%%%%%%%%%%%%%%%%%%%%%%%%%%%%%%%%%%%%%%%%%%%%%%%%%%%%%%%%%%%%%%%%%%%%%%%
%% \section*{Acknowledgment}

%% We thank Kenji Kaneko and Hiroshi Kaminaga of CNRS-AIST JRL for robot hardware support.

%%%%%%%%%%%%%%%%%%%%%%%%%%%%%%%%%%%%%%%%%%%%%%%%%%%%%%%%%%%%%%%%%%%%%%%%%%%%%%%%
\bibliographystyle{IEEEtran}
\bibliography{main.bib}

\end{document}